\documentclass[10pt,twocolumn,letterpaper]{article}

\usepackage{cvpr}
\usepackage{times}
\usepackage{epsfig}
\usepackage{graphicx}
\usepackage{amsmath}
\usepackage{amssymb}
\usepackage{multirow}
\usepackage{subcaption}
\usepackage{verbatim}
\usepackage{tikz}
\usetikzlibrary{calc}
\usepackage{multirow}
\usepackage{colortbl}
\usepackage{subcaption}

\newcommand{\C}{{\cal C}}
\renewcommand{\P}{{\cal P}}

\newcommand{\y}{{\bf y}}

\newcommand{\p}{{\bf p}}
\newcommand{\f}{{\bf f}}
\newcommand{\La}{{\mathcal L}}

\newcommand{\x}{{\bf x}}
\newcommand{\loc}{{\bf d}}
\newcommand{\subsmo}{\text{smo}}
\newcommand{\subcnn}{\text{cnn}}
\newcommand{\subicc}{\text{icc}}

\DeclareMathOperator*{\argmin}{\arg\!\min}

\newcommand{\bfp}{{\bf p}}

\newcommand{\by}{\mathbf{y}}

\newcommand{\be}{\mathbf{e}}

\def\be {\begin{equation}}
\def\ee {\end{equation}}
\def\beas {\begin{eqnarray*}}
\def\eeas {\end{eqnarray*}}
\def\bea {\begin{eqnarray}}
\def\eea {\end{eqnarray}}

\makeatletter
\usepackage{xspace}
\def\@onedot{\ifx\@let@token.\else.\null\fi\xspace}
\DeclareRobustCommand\onedot{\futurelet\@let@token\@onedot}
\newcommand{\figref}[1]{Fig\onedot~\ref{#1}}
\newcommand{\equref}[1]{Eq\onedot~\eqref{#1}}

\makeatother

\definecolor{orange}{rgb}{1,0.9,0.65}
\definecolor{gr}{rgb}{0.9,1,0.6}
\definecolor{bl}{rgb}{0.9,0.8,1}
\definecolor{bg}{rgb}{0.8,0.9,1}
\definecolor{orr}{rgb}{1,0.96,0.75}
\definecolor{grr}{rgb}{0.8,1,0.5}
\definecolor{blr}{rgb}{0.97,0.8,1}
\definecolor{bgr}{rgb}{0.5,0.9,1}
\definecolor{gry}{rgb}{0.92,0.92,0.92}

\newlength\savedwidth

\newcommand{\Ziyu}[1]{{\color{blue}{#1}}}

\usepackage[pagebackref=true,breaklinks=true,letterpaper=true,colorlinks,bookmarks=false]{hyperref}

\cvprfinalcopy 

\def\cvprPaperID{1431} 

\ifcvprfinal\fi
\begin{document}

\title{Instance-Level Segmentation for Autonomous Driving\\ with Deep Densely Connected MRFs}

\author{Ziyu Zhang\ \ \ \ \ \ \ \ Sanja Fidler\ \ \ \ \ \ \ \ Raquel Urtasun\\
Department of Computer Science, University of Toronto\\
{\tt\small \{zzhang, fidler, urtasun\}@cs.toronto.edu}
}

\maketitle
\thispagestyle{empty}

 \begin{abstract}
 \vspace{-2mm}
Our aim is to provide a pixel-wise instance-level labeling of a monocular image in the context of autonomous driving. We build on recent work~\cite{ZhangICCV15} that trained a convolutional neural net to predict instance labeling in local image patches, extracted exhaustively in a stride from an image.  A simple Markov random field model using several heuristics was then proposed in~\cite{ZhangICCV15} to derive a globally consistent instance labeling of the image. In this paper, we formulate the global labeling problem with a novel densely connected Markov random field  and show how to encode various intuitive potentials in a way that is amenable to  efficient mean field inference \cite{krahenbuhl2011efficient}. Our potentials encode the compatibility between the global labeling and the patch-level predictions, contrast-sensitive smoothness as well as the fact that separate regions form different instances. Our experiments on the challenging KITTI benchmark~\cite{kitti} demonstrate that our method achieves a significant performance boost over the baseline~\cite{ZhangICCV15}. 
\end{abstract}

\vspace{-3mm}
\section{Introduction}

Object detection is one of the fundamental open  problems in computer vision. 
The main objective  is to place tight bounding boxes  around each object of interest.
In the past two years, detection performance has almost doubled thanks  to the availability of large datasets as they enable training very deep representations~\cite{alexnet,verydeep}. 
While detection might have been a good proxy when  performance was low, 
recent work has been trying to go beyond  simple boxes by providing a detailed segmentation mask for each object instance~\cite{HariharanSimultaneousECCV2014,yang2012layered,tigheCVPR14,HeCVPR2014,WangCVPR15}. 

\begin{figure}[t!]
\vspace{-2mm}
	\centering
	\includegraphics[width=\linewidth]{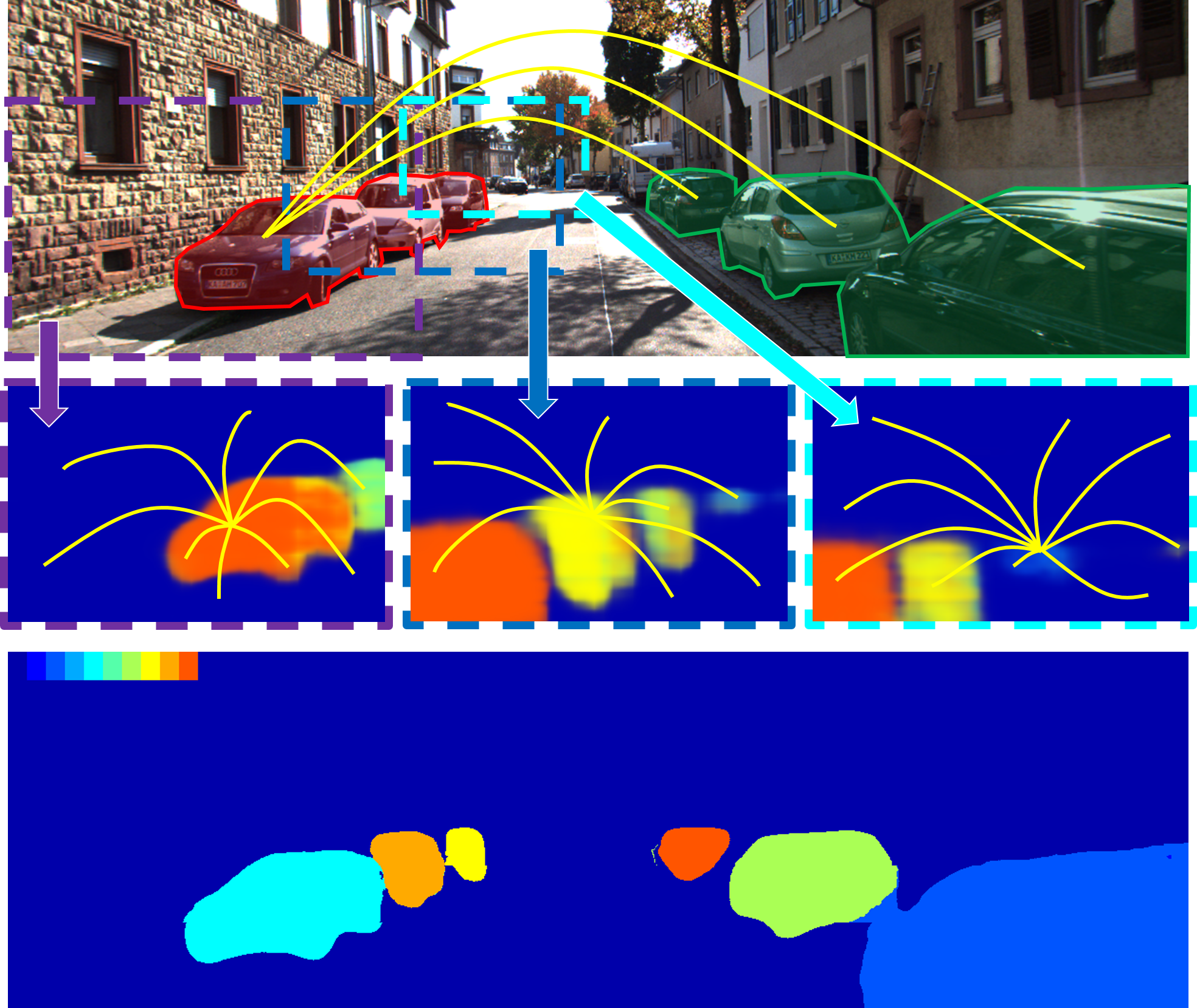}
	\vspace{-6mm}
	\caption{\small Our approach densely samples patches of different sizes from the image (row 1) and exploits a CNN to provide a soft instance labeling of each patch (row 2). Our MRF connects all pixel pairs inside the patches (yellow curves in row 2), as well as all pixel pairs from far away connected components obtained from patch-level CNN predictions (yellow curves in row 1), to provide a globally consistent instance labeling of the image (row 3). }
	\label{fig:intro}
	\vspace{-3mm}
\end{figure}

A mask is in many ways richer than a box: it allows an informed reasoning about occlusion and  depth layering. For robotics applications where depth is available, it further enables a more precise 3D localization and segmentation~\cite{guptaECCV14} which is important for, e.g., obstacle avoidance, route planning and object grasping. An instance mask is also more informative than pixel-wise class labeling as  it allows counting,  important for applications such as retrieval~\cite{LinCVPR14}.


Instance segmentation has been addressed in a variety of ways. In interactive segmentation, approaches like Grabcut~\cite{Rother2004SIGGRAPH,Boykov2004PAMI} require a user-supplied box, or a scribble on the foreground and background, in order to segment the objects. This is typically done with graph-cuts by combining appearance cues and  smoothness. The most common approach to instance segmentation has been to utilize object detections and top-down shape priors to label pixels inside each detected box~\cite{Kumar05,girshick2013rich}. Methods that jointly reason about instance labeling and, possibly,  depth ordering given object detections and class-level semantic segmentation have shown improved performance~\cite{tigheCVPR14}. Recently, approaches that train CNNs to predict instance labeling inside densely sampled image patches have shown very promising performance~\cite{ZhangICCV15}. However, deriving a globally consistent instance labeling of the image from local predictions is a challenging open problem.

In this paper, our goal is to estimate an accurate pixel-level labeling  of  object instances from a single  monocular image in the context of autonomous driving. 
We propose a method that combines the soft predictions of a neural net run on many overlapping patches into a consistent global labeling of the entire image.
We formulate this problem as a densely connected Markov random field (MRF) with several potentials encoding consistency with local patches, contrast-sensitive smoothness as well as the fact that separate regions form different instances. An overview of our MRF is given in Fig. \ref{fig:intro}. Our main technical contribution is a formulation that encodes all potentials in a way that is amenable to  efficient mean field inference \cite{krahenbuhl2011efficient}, and we go beyond \cite{krahenbuhl2011efficient} by including Potts potentials as well. 
Our experimental evaluation shows significant improvements over~\cite{ZhangICCV15} on the challenging KITTI benchmark~\cite{kitti}.

\section{Related Work}

We focus our review on techniques operating on a single monocular image, and divide them into different types.

\vspace{-0.4cm}

\paragraph{Instance-Level Segmentation by Detection.} The most common approach to object instance segmentation is to first  localize objects with a set of bounding boxes, and then exploit top-down information such as  object's shape and appearance in order to accurately segment the object within each box~\cite{Kumar05}.
~\cite{ladicky10,whole} proposed an MRF model  to  identify which detections are true positive, and output  pixel-wise class labels, thus performing instance segmentation.  ~\cite{HaykoECCV12} votes for object centers and uses a MRF to infer the assignment of pixels to object centers. 
R-CNN~\cite{girshick2013rich} was used in~\cite{HariharanSimultaneousECCV2014} to first generate object proposals, and  then predictions from two CNNs (box and region-based) were fused to segment the object inside each box. This idea was extended to RGB-D in~\cite{guptaECCV14}.

In~\cite{yang2012layered}, the authors propose a generative model that takes as input candidate detections and jointly assigns a pixel to an object instance as well as a depth layer. Inference is performed with coordinate ascent  iterating between updating the layer assignment and the parameters of the appearance models. In contrast, we leverage the efficient inference algorithm for Gaussian MRFs~\cite{krahenbuhl2011efficient} to work directly on the pixel level (rather than layers), thus allowing more freedom in the final label assignment.
In~\cite{tigheCVPR14}, semantic segmentation and  object detection are run as a first step. The method then solves for instance labeling and depth ordering by minimizing an integer quadratic program. In our approach we use a CNN trained to directly predict  instance labeling in a stride of local patches, and then solve for a consistent instance labeling  using our densely connected MRF, thus not requiring to explicitly perform object detection. 

\vspace{-0.4cm}

\paragraph{3D CAD Models.} Another line of work matches CAD models  to images~\cite{Aubry14,LimICCV13,guptaCVPR15a}. An image-aligned CAD model effectively provides an instance labeling of the image. CAD matching is however typically slow, and not very robust, as there is a large difference between the appearance of the synthetic models and objects in real images. Instead of matching CAD models,~\cite{Isola13} retrieves object segments from a dataset of labeled objects. Their probabilistic model then aims to find segments that optimally transform into the given image by respecting typical 3D relations. The output is an instance labeling of the image (a ``scene collage'').

\vspace{-0.4cm}

\paragraph{Interactive Segmentation.} Instance-level segmentation has also been done without prior knowledge about how the object looks like. In this line of work, the techniques rely on a user-supplied box, or a scribble on the foreground and background, and then derive the pixel-wise labeling of the object instance. For example, GrabCut~\cite{Rother2004SIGGRAPH}  utilizes annotations in the form of 2D bounding boxes, and computes the  foreground/background models  using EM. ~\cite{Boykov2001ICCV} relies on scribbles as seeds  to
model  appearance of foreground/background, and  uses graph-cuts by combining appearance cues and a  smoothness term~\cite{Boykov2004PAMI}.

\vspace{-0.4cm}

\paragraph{Instances without Object Detection.} Recent work has tried to explicitly reason about instance segmentation (no class detectors need to be run in advance).~\cite{SilbermanECCV14} makes an optimal cut in a hierarchical segmentation tree to obtain object instance regions.~\cite{Liang15} trained a multi-output CNN that jointly predicts pixel-level class labeling of the image as well as bounding box locations and object instance numbers. Off-the-shelf clustering is used to derive the final object instance labeling of the image. In our work, we exhaustively sample bounding boxes and softly score each pixel belonging to a particular object instance. Our main efforts are then devoted to ``clustering'' (merging the predictions) which in our work is done via a densely connected MRF. Parallel to our work,~\cite{torr16} proposes a recurrent neural net to label object instances sequential by keeping a memory of which pixels have been labeled so far. 

We build on~\cite{ZhangICCV15} which trains a CNN on local patches to obtain a depth-ordered pixel-wise instance labeling of each patch. \cite{ZhangICCV15} then uses a MRF along with a connected component algorithm to merge predictions in the possibly overlapping patches into a global instance labeling of the image. In our paper, we propose a densely connected MRF that exploits fast inference~\cite{krahenbuhl2011efficient}, and provides significantly better segmentations due to the dense connectivity in the model.

\section{Object Instance Labeling}

The goal of this paper is to perform instance-level segmentation given a single monocular image. 
We follow~\cite{ZhangICCV15} and learn deep representations to perform this task. 
Our contribution is then a novel densely connected Markov random field that is able to produce a single coherent explanation of the full image and amenable to the efficient mean field inference algorithm \cite{krahenbuhl2011efficient}.  As shown in our experimental evaluation the estimates provided by our approach are significantly better than those of~\cite{ZhangICCV15} in most metrics. 

\subsection{Deep Learning for Instance-level Segmentation}

We follow~\cite{ZhangICCV15} to both generate surrogate ground truth and train the CNN.
We provide the details for completeness. We generate training examples by extracting overlapping patches  at multiple resolutions.
Since KITTI does not have instance-level segmentations, we use~\cite{ChenCVPR14,cadmodel} to obtain the segmentations for our training set. 
We label the instances according to their  depth  within the patch. Thus instances farther away from the camera will get higher IDs.
Using depth ordering is important to produce a single labeling, breaking the symmetry of permutations of instance-level segmentation. E.g., two instances can be labeled as either (1,2) or (2,1). 
We then train a CNN to output a pixel-level instance labeling inside each patch. We use the 
architecture from~\cite{SchwingTR2015}
pre-trained on ImageNet and fine-tune it for instance-level segmentation using our surrogate ground truth. 
The CNN gives us  (probabilistic) pixel-level predictions of instances at the patch level. We propose a model to merge all the local predictions and produce a globally consistent image labeling. This is the  contribution of our paper.


\subsection{Densely Connected Pixel-wise MRF}

Given an image $\x$, we index the image patches with $z$. Let  $\P_z$ be the set of pixels in patch $z$. 
Let $\p_{z,i}$ be the output of the softmax for the $i$-th pixel when we apply the  CNN  to patch $z$. Note that the CNN predicts up to 5 instances as well as background. Thus $\p_{z,i}$ is a 6-D vector.
The goal is to merge all the patch-level information and come up with a single explanation of the image in terms of all instances. We restrict the maximum number of instances to be 9 per image, which is sufficient for KITTI. 
Thus our global label space $\La_\text{g}$ is $\{0,1, \cdots, 9\}$ with $0$ encoding background.
Let $\by$ be the labeling of each pixel in the image with $y_i \in \La_\text{g}$. 
Unlike \cite{ZhangICCV15}, we are not interested in ordering the instances by depth. Thus any labeling that separates instances is valid.


We propose a novel densely connected pixel-wise MRF to solve for the problem of labeling the full image given the local patch-based predictions. 
The corresponding Gibbs energy $E(\y)$ of our MRF consists of three main terms: a pairwise \textit{smoothness} term, a pairwise \textit{local CNN prediction} term and a pairwise \textit{inter-connected component} term, each encoding different intuitions about the problem: 
\begin{equation}
E(\y)  = E_\subsmo(\y) + E_\subcnn(\y) + E_\subicc(\y).
\end{equation}
Note that all terms are defined over densely connected pixel pairs. We cannot use the  CNN output as a unary potential, as the label space of the local patches and the global image are different, i.e., only 6 labels (including background) possible locally, and instance 2 in a local patch might be totally different from instance 2 in another patch far away. 

We now describe each term in more details.

\vspace{-1mm}
\subsubsection{Smoothness: $E_\subsmo(\y)$}

Following~\cite{krahenbuhl2011efficient}, we incorporate a contrast-sensitive smoothness term into our MRF to remove noisy tiny regions. The idea is to describe each pixel with a feature vector, and define a potential that encourages pixels with similar features to be more likely assigned the same label. The typical feature for each pixel has been color and position on image. 

In our problem we use the combination of position and the output of the CNN to form our feature space. Our CNN is trained to differentiate between object instances, so the probability vectors that the CNN outputs are a very strong cue of how likely two pixels  belong to the same object. 
Further, we use the position feature so that the smoothness has a lower influence between far apart regions in order not to over-smooth the result.
Notice that we do not use color as a feature. This is because different object instances can take similar colors, and color may be somewhat deceiving due to shadows, saturation and specularities.

Formally, let $\loc_i$ be the 2-D position vector for pixel $i$ in the image. We define the contrast-sensitive smoothness term as a sum of patch-specific contrast-sensitive smoothness terms, each defined over all pixel pairs in the patch:
\begin{equation}
E_\subsmo(\y) = 
\sum_z \sum_{i, j: i, j \in \P_z, i < j} \varphi_\subsmo^{(z,i,j)}(y_i, y_j), 
\end{equation}
where the potential is defined as 
\begin{equation}
\varphi_\subsmo^{(z,i,j)}(y_i, y_j) = w_\subsmo \mu_\subsmo(y_i,y_j)  k_\subsmo \left(\f_i^{(z)}, \f_j^{(z)}\right).
\end{equation}
Here $w_\subsmo$ is the weight for the potential (which we learn) controlling the degree of smoothness, and $k_\subsmo$ denotes a Gaussian kernel defined as  
\[
k_\subsmo \left(\f_i^{(z)}, \f_j^{(z)}\right) = \exp \left( - \frac{\|\p_{z,i} - \p_{z,j}\|_2^2}{2\theta_1^2} - \frac{\|\loc_i - \loc_j\|_2^2}{2\theta_2^2} \right),
\]
where $\f_i^{(z)}$ contains both the position $\loc_i$ and the output of the CNN $\p_{z,i}$.  
Note that  $\theta_1$ and $\theta_2$ scale the features to reflect our notion of ``closeness" in the feature space.
Finally, 
the compatibility function $\mu_\subsmo(y_i, y_j)$ in the potential takes the form of the Potts model:
\[
\mu_\subsmo(y_i,y_j) = \begin{cases}
1, & \text{if } y_i \neq y_j \\
0, & \text{otherwise} \end{cases}.
\]
This penalizes two pixels with similar positions and CNN predictions to have different labels. 


 \begin{figure}[t!]
 \vspace{-1mm}
	\centering
	\includegraphics[width=\linewidth]{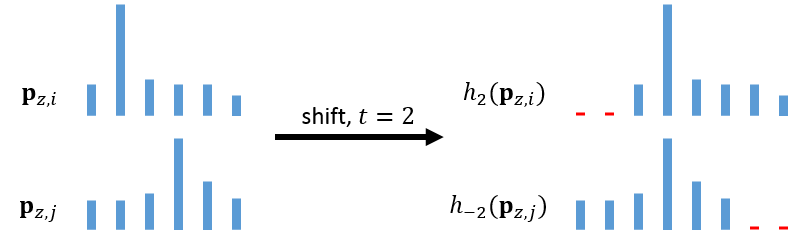}
	\caption{\small Function $h_2(\cdot)$ {\bf prepends} 2 zeros to vector $\bfp_{z,i}$ while $h_{-2}(\cdot)$ {\bf appends} 2 zeros to vector $\bfp_{z,j}$. This is equivalent to shifting $\bfp_{z,i}$ by 2 units towards right and zero-pad both vectors to make them aligned. The shift matches the modes of $\bfp_{z,i}$ and $\bfp_{z,j}$. We input this shifted vector pair into our Gaussian kernel.}
	\label{fig:shift}
	\vspace{-2mm}
\end{figure}

\subsubsection{Local CNN Prediction: $E_\subcnn(\y)$}

Any given patch contains only a subset of the instances present in the full image. 
When we produce our image-level labeling, we would like to maintain the separation into different instances estimated at the local level (via the CNN), while producing a coherent labeling across all patches. 
For example, if the CNN predicts that in patch $z$ pixel $i$ belongs to instance 1 while pixel $j$ belongs to instance 2, then any global configuration with $y_i \neq y_j$ should be encouraged. However, it turns out to be important to have some preference over the ordering just to break the symmetry in our model to kick off the inference algorithm. We thus encourage ordering in the global labeling (in this example $y_i < y_j$). 

To encode this patch-image compatibility in our energy, we define the local CNN prediction term
as a sum of patch-specific compatibility terms, each defined over all pixel pairs in the patch:
\begin{equation}
E_\subcnn(\y) = \sum_z \sum_{i, j: i, j \in \P_z, i < j} \varphi_\subcnn^{(z,i,j)}(y_i, y_j).
\end{equation}
The potential $\varphi_\subcnn^{(z,i,j)}(y_i, y_j)$ should ideally encode the fact that we want global instance labeling to agree with local predictions. That is, if two pixels are likely to be of the same (different) instance at the local level, they should also be the same (different) at the global level. 
This could be simply encoded with a compatibility potential that $y_i$ and $y_j$ are encouraged to have the same label if the output of the CNN $\p_{z,i}$ and $\p_{z,j}$ are similar, and their relative ordering ($y_i>y_j$ or vice versa) is respected if the CNN predicts them to be of different instances. 
This  naive approach, however,  will break the efficiency of inference, as we will no longer be able to use Gaussian filtering.  Gaussian filtering is however crucial, since our local CNN prediction term is fully connected at patch level.

Instead, one  of the main contributions of our paper is to approximate such compatibility potentials as a series of Gaussian potentials. Each potential is composed of a Gaussian kernel applied to a shifted version of the local softmax probabilities:
\begin{equation}
\varphi_\subcnn^{(z,i,j)}(y_i,y_j) = \sum_{t = -T}^{T} \varphi_\subcnn^{(z,t,i,j)}(y_i,y_j),
\end{equation}
where $T$ is the maximum shift allowed (fixed to be 2 in our experiments). 

We define the shifted pairwise potential $\varphi_\subcnn^{(z,t,i,j)}(y_i,y_j)$ as a product of its weight, a compatibility function and a Gaussian kernel defined over pairs of shifted local CNN predictions:
\begin{equation}
\varphi_\subcnn^{(z,t,i,j)}(y_i,y_j) = w_\subcnn^{(s(z))} \mu_\subcnn^{(t)}(y_i, y_j) k_\subcnn^{(t)}(h_t(\bfp_{z,i}), h_{-t}(\bfp_{z,j})), \label{eq:phi_cnn}
\end{equation}
where $w_\subcnn^{(s(z))}$ is the weight which depends on the size $s(z)$ of patch $z$, and $k_\subcnn^{(t)}$ is a Gaussian kernel characterized by its precision matrix $\Lambda_\subcnn^{(t)}$. 

When $t>0$, a shift towards right is applied to $\p_{z,i}$ to create $h_t(\p_{z,i})$ while a shift towards left is applied to $\p_{z,j}$ to create $h_{-t}(\p_{z,j})$. 
Note  that shifting by $t$ requires {\bf prepending} $t$ zeros, while shifting by $-t$ requires {\bf appending} $t$ zeros. We refer the reader to \figref{fig:shift} for a visualization of this idea. 
If the modes of $\p_{z,i}$ and $\p_{z,j}$ match for any positive $t$, it means that the label of pixel $i$ is predicted to be smaller than pixel $j$ in patch $z$ by the CNN. This is the case shown in~\figref{fig:shift}. Therefore, globally we prefer any configuration  with $y_i < y_j$. The reverse is also true that if we achieve a good match with a negative $t$, then we prefer any configuration with $y_i>y_j$. If the best match is achieved without shift, it means that we prefer $y_i=y_j$. 
This can be encoded via the following compatibility function:
\begin{equation}
\mu_\subcnn^{(t)}(y_i, y_j) = \begin{cases}
-1, & \text{if } y_i < y_j, t > 0 \\
-1, & \text{if } y_i > y_j, t < 0 \\
-1, & \text{if } y_i = y_j, t = 0 \\
0, & \text{otherwise}
\end{cases}.
\end{equation}
Note that a negative value in $\mu_\subcnn^{(t)}(y_i, y_j)$ implies that we encourage the configuration. 

\begin{figure*}[htb!]
\vspace{-5mm}
\centering

\begin{subfigure}[b]{0.28\linewidth}
\includegraphics[height=1.67in,width=\linewidth]{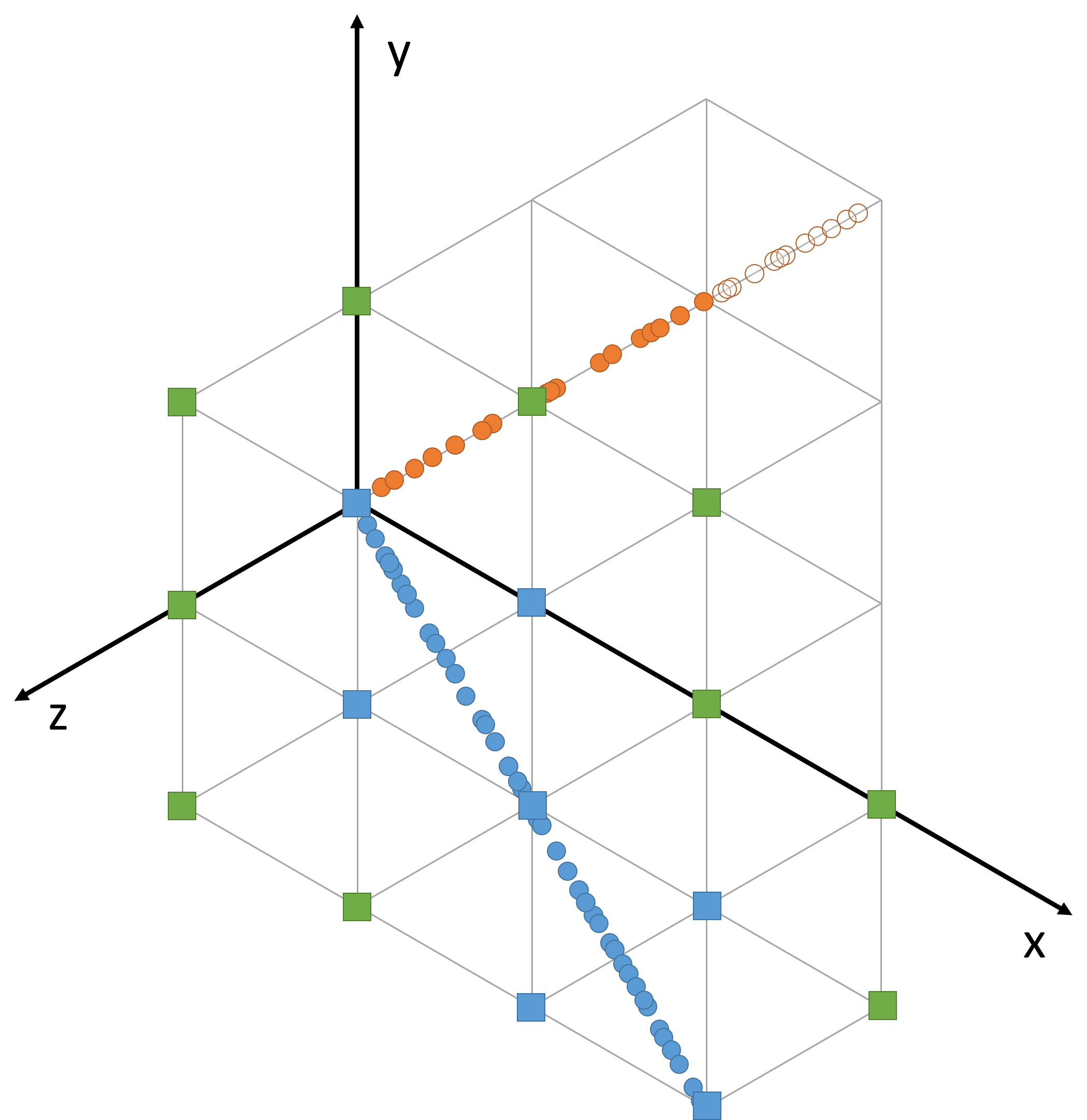}
\caption{}
\label{fig:lattice}
\end{subfigure}
~
\begin{subfigure}[b]{0.65\linewidth}
\includegraphics[height=1.67in,width=\linewidth]{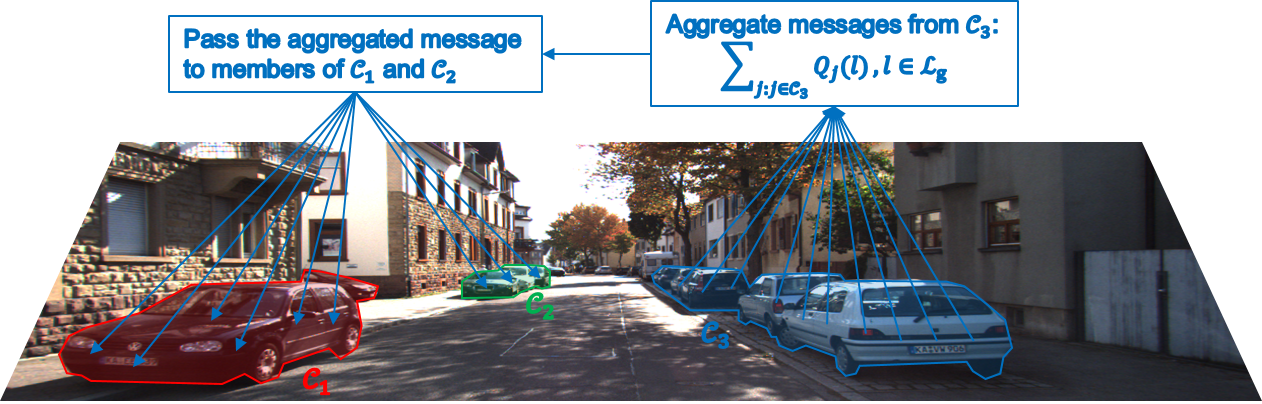}
\caption{}
\label{fig:inter-cco}
\end{subfigure}
\vspace{-3.0mm}
\caption{\small {\bf (a)} Suppose that $\{\p_{z,i}\}$ are positive scalars. When $t=1$, both $h_t(\bfp_{z,i})$ and $h_{-t}(\bfp_{z,j})$ are 2-D. The corresponding permutohedral lattice (in gray) lies on a 2-D hyperplane in a 3-D space. $\{h_{-t}(\bfp_{z,j})\}$ are embedded in the hyperplane ({\color{blue}blue dots}). The embeddings follow a 1-D subspace due to zero-appending. Value of $\{Q_j(l')\}$ splats onto the vertices of their respective enclosing simplexes ({\color{blue}blue squares}). After Gaussian blurring, vertices shown as {\color{green}green squares} also get non-zero value. An extra step is to embed $\{h_t(\bfp_{z,i})\}$  ({\color{red}red solid dots} and {\color{red}red open dots}). The embeddings lie on another 1-D subspace due to zero-prepending. Finally, convolution is evaluated at the new embeddings. Only {\color{red}red solid dots} get non-zero values, while {\color{red}red open dots} get a zero due to the inactivity of the vertices of their enclosing simplex. {\bf (b)} Three car blobs are shown in the image. During each mean field update, we sum up individual messages from  members of $\C_3$ and pass the summation to each member of $\C_1$ and $\C_2$. Similarly for messages from $\C_1$ to $\C_2$ and $\C_3$, and messages from $\C_2$ to $\C_1$ and $\C_3$.}
\vspace{-1.5mm}
\end{figure*}

\vspace{-2mm}
\subsubsection{Inter-Connected Component: $E_\subicc(\y)$}

So far our MRF encourages smoothness as well as that global instance labeling to agree with local predictions.  However, nothing prevents instances that are far apart and thus do not appear together in any patch from having the same label.
Towards this goal, for each  pixel $i$, we compute the probability that it belongs to foreground, by summing the output of local CNN predictions and re-normalizing. 
By thresholding this probability, we obtain a binary mask of activation. We index each connected component of the foreground of the binary mask with $m$, and the pixels it contains with $\C_m$.
Each component might contain more than one car.
However, it is reasonable to assume that each instance will never appear in two different components. 
We encode this in the inter-connected component term as a sum of terms defined over component pairs, and each of the terms fully connects cross-component pixel pairs:
\begin{equation}
	E_\subicc(y) = \sum_{m,n: m<n} \sum_{i,j: i \in \C_m, j \in \C_n} w_\subicc \mu_\subicc(y_i, y_j),
\end{equation}
with $w_\subicc$ the weight and $\mu_\subicc(y_i, y_j)$ a Potts potential
\begin{equation}
\mu_\subicc(y_i, y_j) = \begin{cases}
1, & \text{if } y_i = y_j \\
0, & \text{otherwise}
\end{cases}.
\end{equation}
While this potential is not Gaussian, and we have dense connections, in the next section we show that the updates can still be computed in linear time.


\subsection{Efficient Inference}

Inference in our model consists of estimating the minimum energy configuration
\[
\y^* = \argmin_{\y} E_\subsmo(\y) + E_\subcnn(\y) + E_\subicc(\y).
\]
Unfortunately this is NP-hard. 
Instead, we perform efficient approximated inference via mean field. Towards this goal, we approximate the Gibbs distribution $P(\y|\x)=\frac{1}{Z(\x)} \exp (-E(\y|\x))$ with a fully decomposable distribution $Q(\y|\x) = \prod_i Q_i(y_i|\x)$. Note that we drop the conditioning from now on to simplify notation. 

Mean field computes updates by iteratively minimizing the KL-divergence between the approximated distribution $Q(\y)$ and the true  distribution $P(\y)$. 
We use  an iterative algorithm which updates the local distributions $\{Q_i(y_i)\}$ in parallel. 
In our model, the updates can be derived as
\begin{align}
& \log Q_i(y_i=l) \nonumber \\
= & - \sum_{z:i\in\P_z} \sum_{l': l'\in\La_\text{g}} \sum_{j: j \in \P_z, j \neq i} \varphi_\subsmo^{(z,i,j)}(l, l') Q_j(l') \label{update:smooth} \\
& - \sum_{z: i \in \P_z} \sum_{t = -T}^T \sum_{l': l'\in\La_\text{g}} \sum_{j: j \in \P_z, j \neq i} \varphi_\subcnn^{(z,t,i,j)}(l,l') Q_j(l') \label{update:prediction} \\
& - w_\subicc \sum_{n: n \neq m, i \in \C_m} \sum_{j: j\in\C_n} Q_j(l) - \log(Z_i), \label{update:inter-cco}
\end{align}
with $Z_i$ the local partition function which is easily computable as it only depends on a single node. 
We refer the reader to suppl. material for the derivation of these updates. 
We use the uniform distribution as our initialization. We now describe how to compute the updates efficiently.

\vspace{-0.4cm}

\paragraph{Smoothness.} \equref{update:smooth} can be computed efficiently using the same high-dimensional Gaussian filtering algorithm of \cite{krahenbuhl2011efficient}. This results in linear updates in the number of pixels. 

\begin{table*}[!htbp]
\vspace{-5mm}
\centering
\small
\addtolength{\tabcolsep}{-1.6pt}
\begin{tabular}{|c | c | c | c | c | c | c | c | c | c | c | c |}
\hline
& & \multicolumn{1}{>{\columncolor{blr}}c|}{{\bf Cls. Eval}} & \multicolumn{9}{>{\columncolor{bgr}}c|}{{\bf Instance Evaluation}} \\
\hline
& & IoU & MWCov & MUCov & AvgPr & AvgRe & AvgFP & AvgFN & InsPr & InsRe & InsF1\\
\hline\hline
\multirow{4}{*}{~\cite{ZhangICCV15}} & ConnComp & 77.1 & 66.7 & 49.1 & 82.0 & 60.3 & 0.465 & 0.903 & 49.1 & 43.0 & 45.8\\
\cline{2-12}
& Unary & {77.6} & 65.0 & 48.4 & 81.7 & 62.1 & 0.389 & 0.688 & 46.6 & 42.0 & 44.2\\
\cline{2-12}
& Unary+LongRange & {77.6} & 66.1 & 49.2 & {82.6} & 62.1 & {0.354} & 0.688 & 48.2 & 43.1 & 45.5\\
\cline{2-12}
& Unary+LR (w/ DeepLab~\cite{chen14semantic}) & {77.7} & 68.2 & 50.2 & {\bf 85.3} & {63.2} & {\bf 0.285} & {\bf 0.562} & 39.5 & 40.1 & 39.8\\
\hline\hline
\multirow{4}{*}{{\bf Ours}} & LocCNNPred & 77.4 & 58.3 & 40.9 & 80.4 & 62.6 & 0.403 & 0.681 & 25.3 & 32.9 & 28.6\\
\cline{2-12}
& LocCNNPred+InterConnComp & 76.8 & 65.7 & 50.3 & 79.9 & {\bf 63.4} & 0.507 & {0.618} & 35.8 & 46.4 & 40.4\\
\cline{2-12}
& Full & 77.1 & {69.3} & {50.6} & 80.5 & 57.7 & 0.451 & 1.076 & {56.3} & {47.4} & {51.5}\\
\cline{2-12}
& Full (w/ DeepLab~\cite{chen14semantic}) & {\bf 78.5} & {\bf 73.7} & {\bf 54.3} & 82.8 & 61.3 & 0.458 & 0.812 & {\bf 63.3} & {\bf 51.6} & {\bf 56.8}\\
\hline
& \multicolumn{11}{>{\columncolor{orr}}c|}{{\bf With Post-processing}} \\
\hline
\multirow{4}{*}{~\cite{ZhangICCV15}} & ConnComp & 77.2 & 66.8 & 49.2 & 81.8 & 60.3 & 0.465 & 0.903 & 49.8 & 43.0 & 46.1\\
\cline{2-12}
& Unary & {77.4} & 66.7 & 49.8 & 81.6 & 61.2 & 0.562 & 0.840 & 44.1 & 44.7 & 44.4\\
\cline{2-12}
& Unary+LongRange & {77.4} & 67.0 & 49.8 & 82.0 & 61.3 & 0.479 & 0.840 & 48.9 & 43.8 & 46.2\\
\cline{2-12}
& Unary+LR (w/ DeepLab~\cite{chen14semantic}) & 77.3 & 70.9 & 52.2 & {\bf 85.7} & {61.7} & 0.597 & 0.736 & 40.2 & 45.9 & 42.8\\
\hline\hline
\multirow{4}{*}{{\bf Ours}} & LocCNNPred & 76.7 & 67.5 & 52.9 & 82.5 & 61.3 & 0.646 & 0.743 & 39.4 & 51.6 & 44.7\\
\cline{2-12}
& LocCNNPred+InterConnComp & 76.3 & 68.1 & {53.9} & 80.7 & {\bf 62.2} & 0.708 & {\bf 0.701} & 42.1 & {52.2} & 46.6\\
\cline{2-12}
& Full & 77.0 & {69.7} & 51.8 & {83.9} & 57.5 & {\bf 0.375} & 1.139 & {65.3} & 50.0 & {56.6}\\
\cline{2-12}
& Full (w/ DeepLab~\cite{chen14semantic}) & {\bf 78.5} & {\bf 74.1} & {\bf 55.2} & 84.7 & 61.3 & {0.417} & 0.833 & {\bf 70.9} & {\bf 53.7} & {\bf 61.1}\\
\hline
\end{tabular}
\vspace{-2mm}
\caption{Instance-level and Class-level Evaluation on the {\bf Test Set} (144 images). See text for the explanation of the metrics. For `AvgFP' and `AvgFN' smaller is better, for the rest higher is better.}
\label{tab:metricsTest}
\end{table*}

\vspace{-0.4cm}

\paragraph{Local CNN Prediction.}
The inner most summation over pixels in~\equref{update:prediction} is the fundamental building block for the computation of the entire term. Explicitly it is given as 
\begin{equation}
w_\subcnn^{(s(z))} \mu_\subcnn^{(t)}(l, l') \sum_{j: j \in \P_z, j \neq i} k_\subcnn^{(t)}(h_t(\bfp_{z,i}), h_{-t}(\bfp_{z,j})) Q_j(l'). \nonumber
\end{equation}
This can be interpreted as a convolution with a Gaussian kernel $G_{\Lambda_\subcnn^{(t)}}$ evaluated at $h_t(\p_{z,i})$. Since Gaussian convolution is essentially a low-pass filter, by the sampling theorem we can convolve a downsampled $\{Q_j(l')\}$ with the Gaussian kernel, and upsample the output to compute convolution at $h_t(\p_{z,i})$. Following~\cite{krahenbuhl2011efficient}, we use the efficient permutohedral lattice data structure~\cite{adams2010fast} to perform downsampling, convolution and upsampling.
In the standard case of Gaussian filtering with permutohedral lattice, we embed a set of features encoding the position of $\{Q_j(l')\}$ in the hyperplane in which the lattice lies. We then downsample by splatting the value of each $Q_j(l')$ onto the vertices of its enclosing simplex with barycentric weights. Then Gaussian blurring is performed over lattice points along each lattice direction. The final result is then computed by gathering values from lattice points for the already embedded features. 
Due to the fact that we introduced different shifts for the two elements of the kernel $h_t(\p_{z,i})$ and $h_{-t}(\p_{z,j})$, apart from the set of features $\{h_{-t}(\p_{z,j})\}$ we need to embed in the first place, we have to embed an extra set of features $\{h_t(\p_{z,i})\}$ at which we evaluate the convolution, in contrast to the standard case. 
An example is in~\figref{fig:lattice}.
As in the standard case, we first embed $\{h_{-t}(\bfp_{z,j})\}$ encoding the position of $\{Q_j(l')\}$ in the hyperplane. Note, however, that the features lie in a subspace of the hyperplane as they are padded with zeros (e.g., a line on a plane in~\figref{fig:lattice}).
We then distribute the value of each $Q_j(l')$ onto the vertices of its enclosing simplex.  
This is followed by the filtering step over the lattice. As an extra step in contrast to the standard case, we now need to embed the first element of the kernel, i.e., $\{h_t(\bfp_{z,i})\}$, in the hyperplane, and the embeddings lie in another subspace. Finally we evaluate the convolution at the new embeddings. 

\vspace{-0.4cm}

\paragraph{Inter-Connected Component.} The first term in~\equref{update:inter-cco} is not a Gaussian kernel, however it is densely connected. This means that potentially we have an update quadratic in the number of pixels. 
We exploit the fact that all members within a connected component have the exact same pairwise interaction with all other pixels not in the component. 
This implies that all members within a connected component receive the exact same messages passed from other components during each update. Note that it is linear to sum up the individual messages within a connected component and pass this summation $\sum_{j: j\in\C_n} Q_j(l)$ to the members of other components. We visualize the message passing procedure for the inter-connected component potential in \figref{fig:inter-cco}.


\vspace{-2mm}
\section{Experimental Evaluation}

We evaluate our approach on the challenging KITTI benchmark~\cite{kitti}. In particular, we use a subset of $3,524$ images from $55$ videos and divide the images into training/validation/test sets such that given a video, all its images are exclusively contained in only one of the three sets. Altogether, we use $3260$ images for training, $120$ for validation, and $144$ for testing. $131$ images from either our validation or test set have been manually annotated with pixel-wise instance labeling for \emph{Cars} by~\cite{ChenCVPR14}. We labeled the rest of the 133 images. Thus all validation and test images have ground truth annotations.
\vspace{-0.4cm}
\paragraph{Implementation Details.} We generate surrogate ground truth for our training images with~\cite{ChenCVPR14} and train our CNN as in~\cite{ZhangICCV15}. In another experiment we also use the architecture DeepLab-LargeFOV from~\cite{chen14semantic}. By changing the CNN architecture from the naive adaptation of VGG-16 by~\cite{SchwingTR2015} to DeepLab-LargeFOV (denoted by `w/ DeepLab~\cite{chen14semantic}' in our results), we observe substantial performance gain for our approach but a slight drop for the baselines. Thus we report results with both architectures.
For our validation/test images (with typical size $375\times1242$), we extract densely overlapping patches of three sizes: large ($270\times432$), medium  ($180\times288$), and small ($120\times192$) in a sliding window fashion. We run the extracted patches through the CNN to obtain local instance predictions.
Following~\cite{krahenbuhl2011efficient}, we apply a pixel-wise normalization to Gaussian kernels. We also normalize the aggregated message of each connected component by the number of pixels it contains. Normalization is able to cancel the bias caused by highly variable instance sizes.
We tune all weights, hyper-parameters and kernel widths in our MRF on the validation set. We fix the number of iterations of the mean field update to be $50$ in all experiments  in the paper.

\vspace{-0.4cm}
\paragraph{Baselines.} We re-train the approach (in three different instantiations) proposed in~\cite{ZhangICCV15} on our validation set to obtain three strong baselines. Note that~\cite{ZhangICCV15} and our method use the exact same CNN unaries and patch extraction method, so we evaluate the two different MRFs on equal footing. The first baseline `ConnComp' is the `connected components ordering' potential in~\cite{ZhangICCV15} which applies a connected component algorithm to heuristically merged object instances and orders them according to their positions along the vertical axis. The second baseline `Unary' additionally adds the `CNN energy' potential in~\cite{ZhangICCV15}. The third baseline `Unary+LongRange' further adds the pairwise `long-range connections' from~\cite{ZhangICCV15}. We found that their pairwise `short-range connections' generally hurts performance, so we do not include it in our baselines.

\begin{figure*}[htb!]
\vspace{-3mm}
\includegraphics[width=0.247\linewidth]{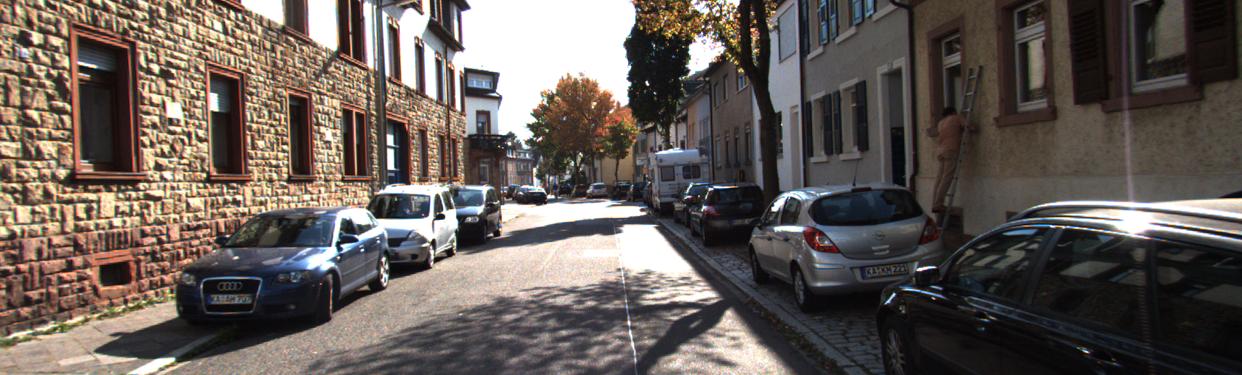}
\includegraphics[width=0.247\linewidth]{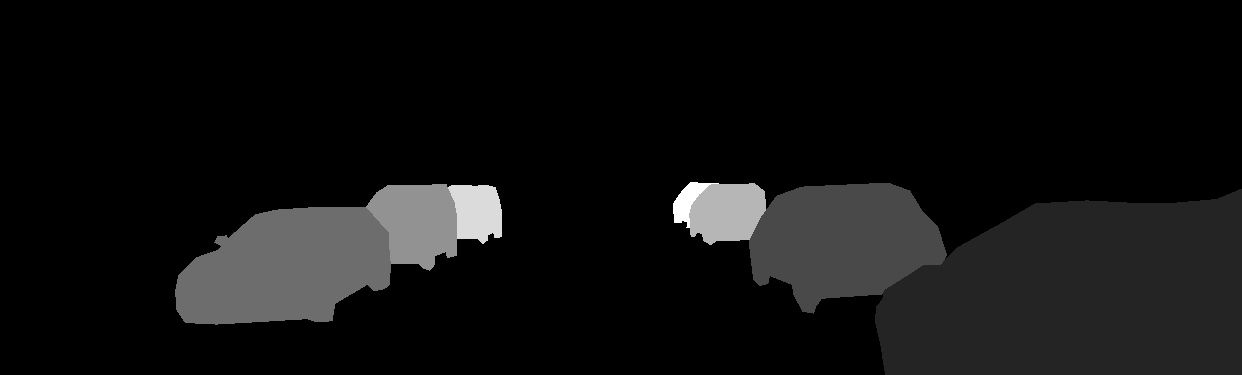}
\includegraphics[width=0.247\linewidth]{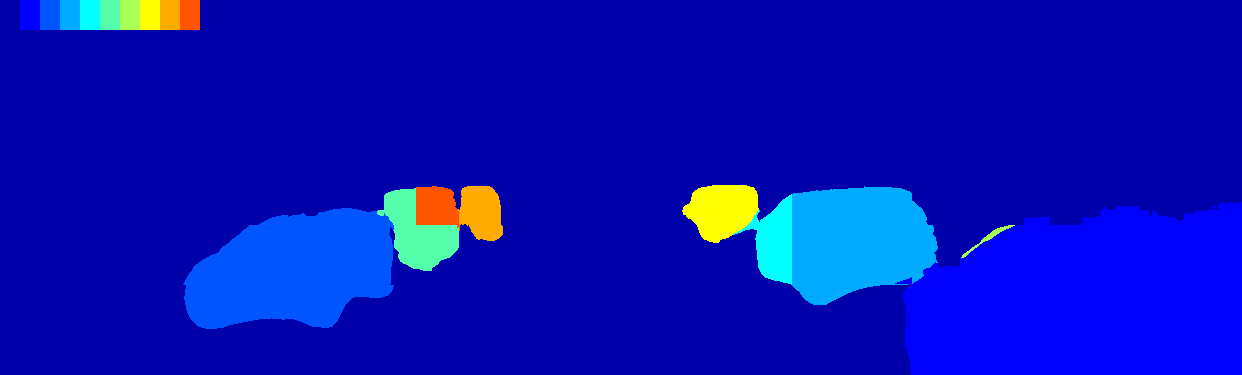}
\includegraphics[width=0.247\linewidth]{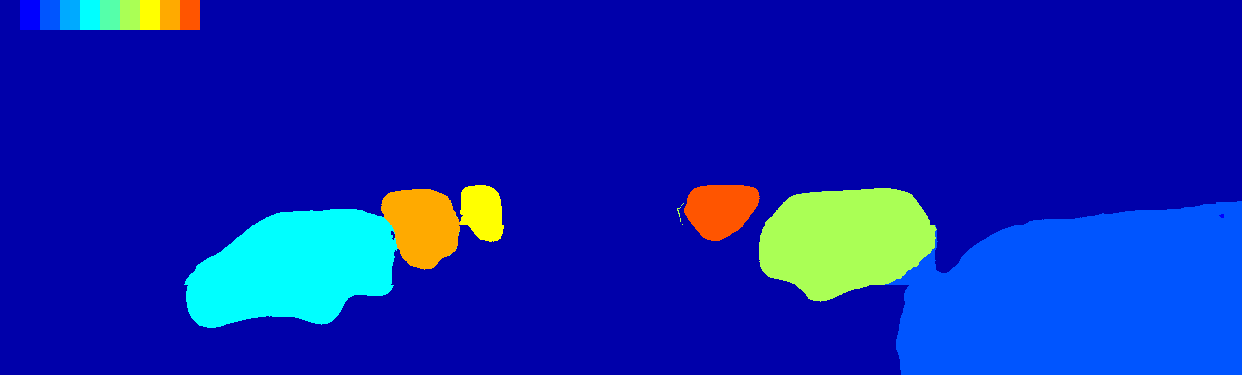}\\[0.3mm]
\includegraphics[width=0.247\linewidth]{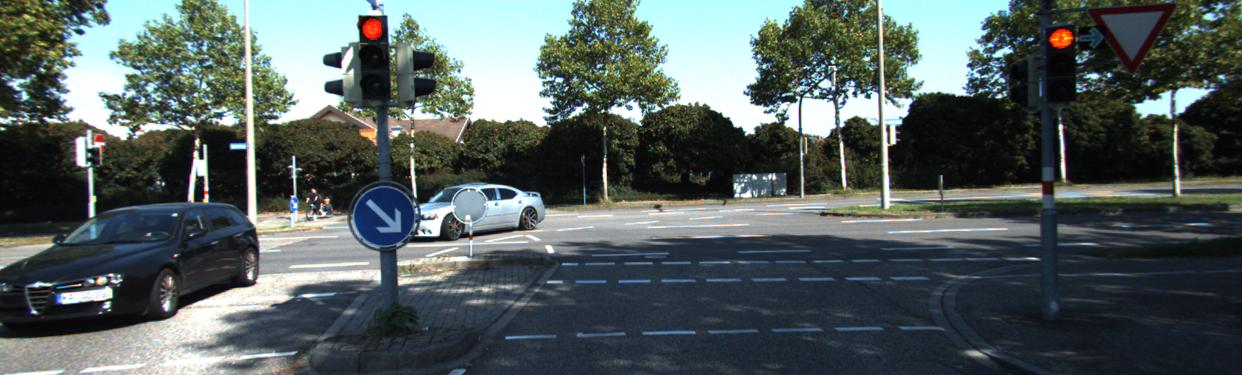}
\includegraphics[width=0.247\linewidth]{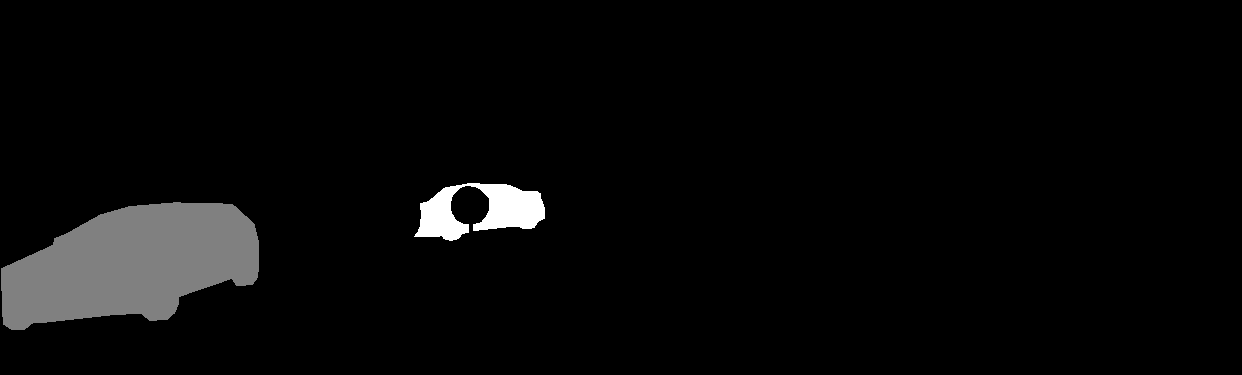}
\includegraphics[width=0.247\linewidth]{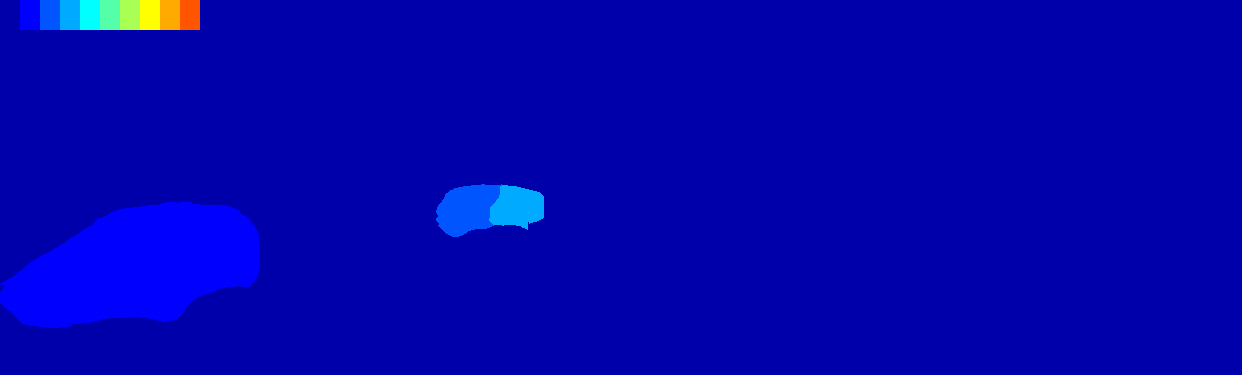}
\includegraphics[width=0.247\linewidth]{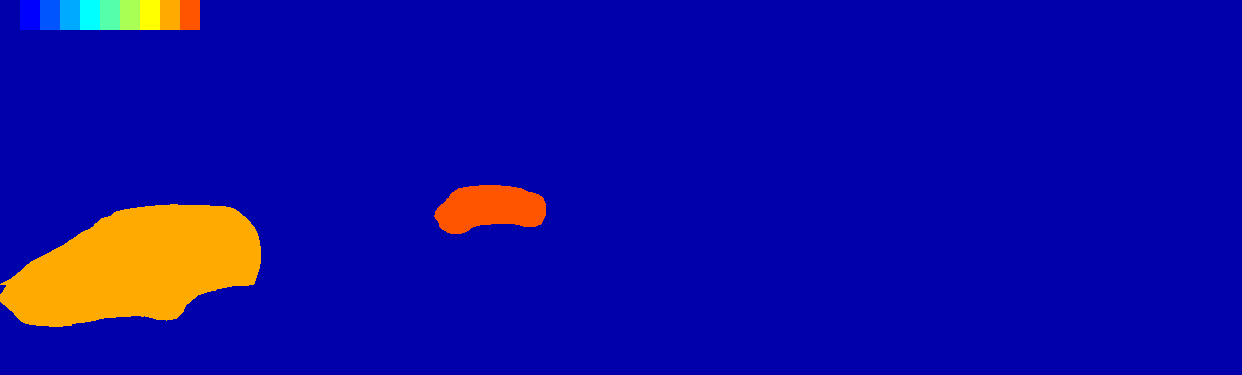}\\[0.3mm]
\includegraphics[width=0.247\linewidth]{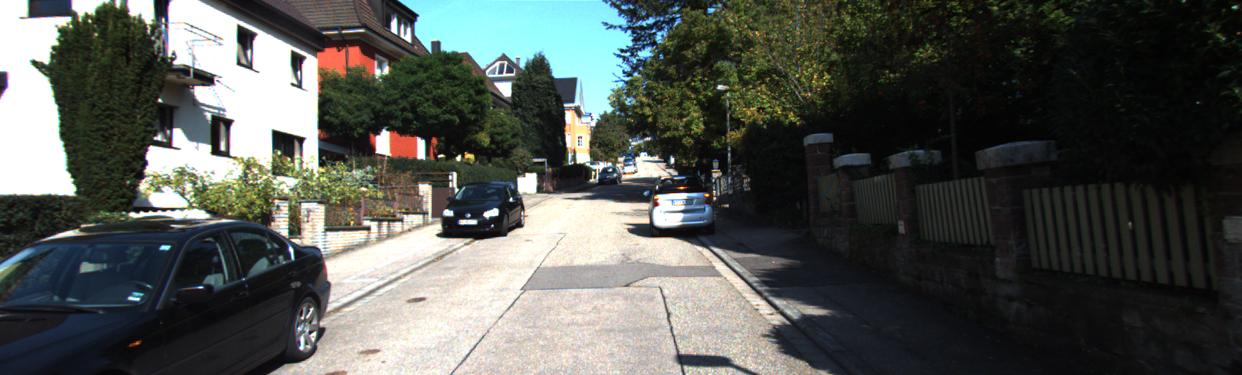}
\includegraphics[width=0.247\linewidth]{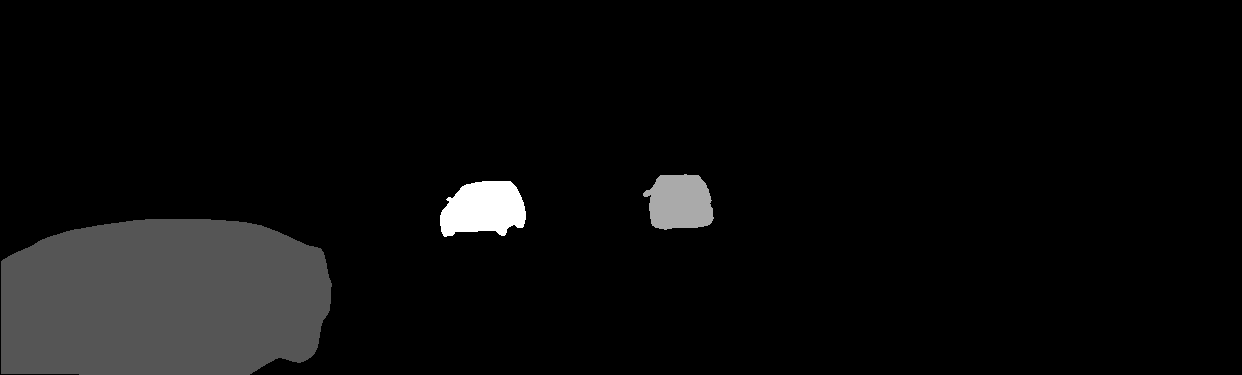}
\includegraphics[width=0.247\linewidth]{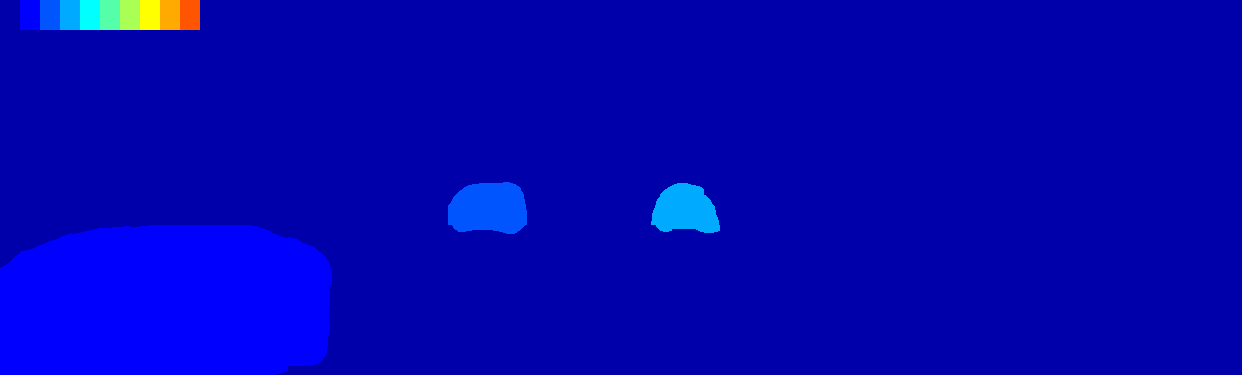}
\includegraphics[width=0.247\linewidth]{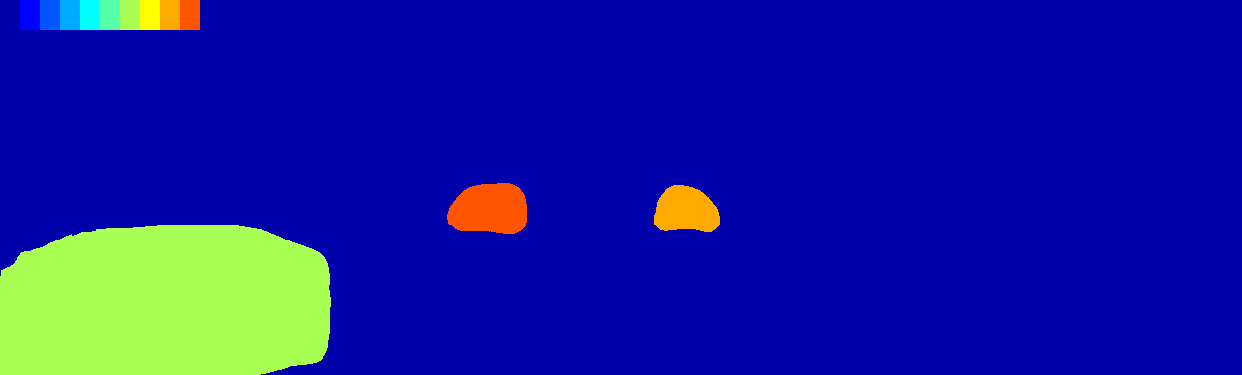}\\[0.3mm]
\includegraphics[width=0.247\linewidth]{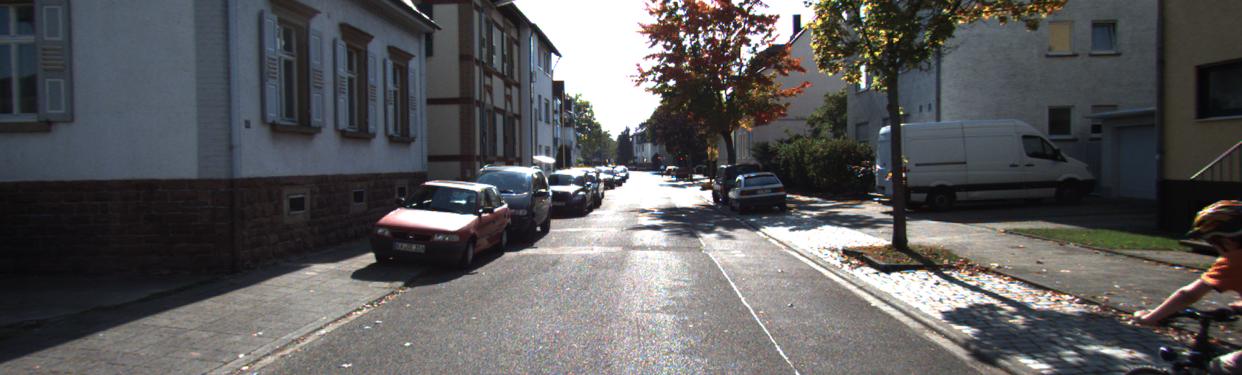}
\includegraphics[width=0.247\linewidth]{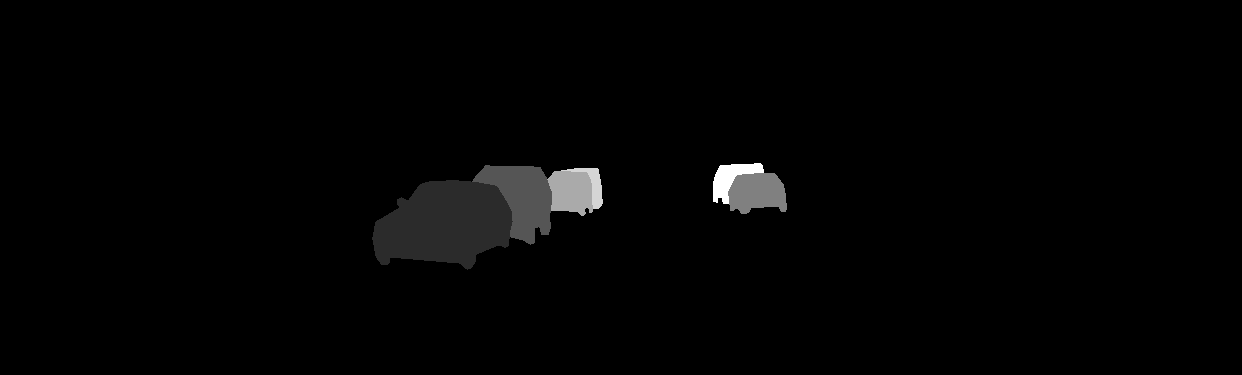}
\includegraphics[width=0.247\linewidth]{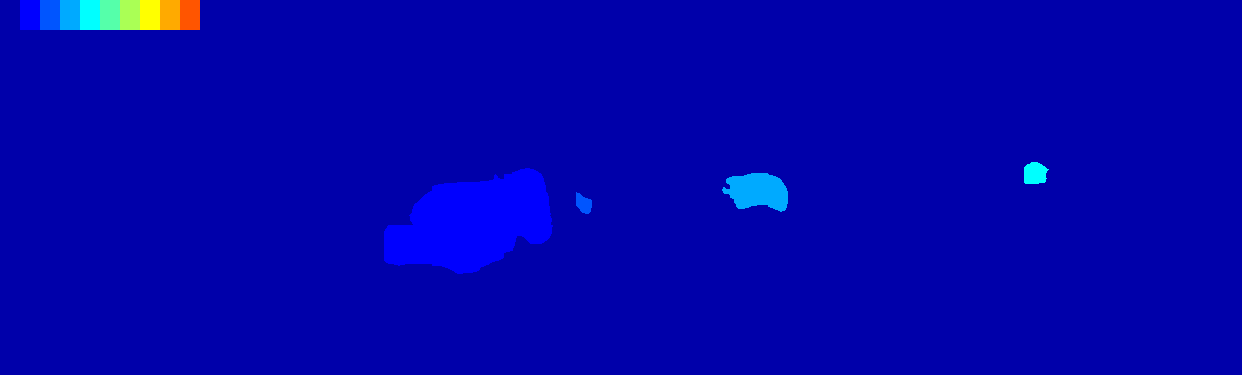}
\includegraphics[width=0.247\linewidth]{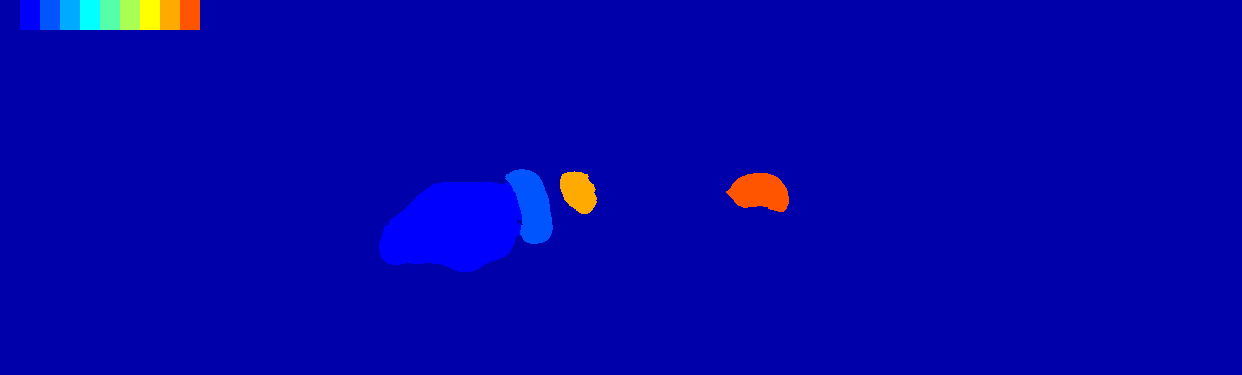}\\[0.3mm]
\includegraphics[width=0.247\linewidth]{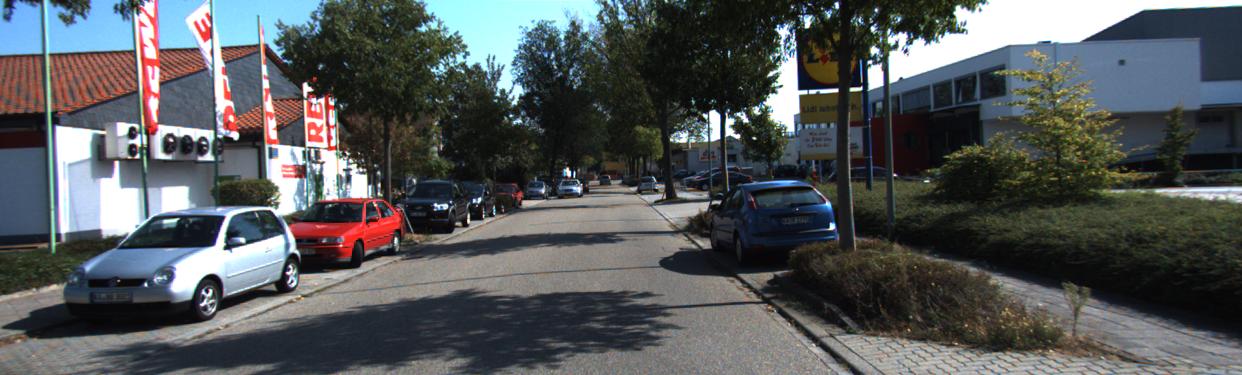}
\includegraphics[width=0.247\linewidth]{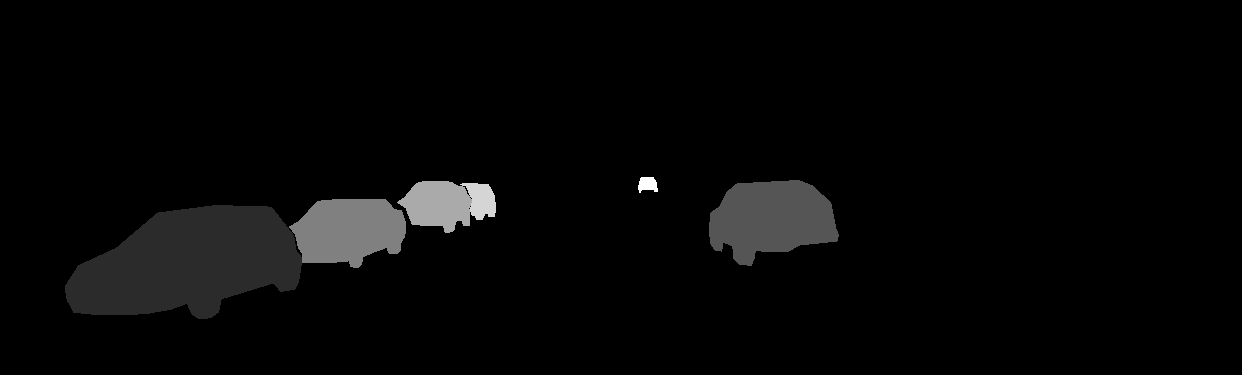}
\includegraphics[width=0.247\linewidth]{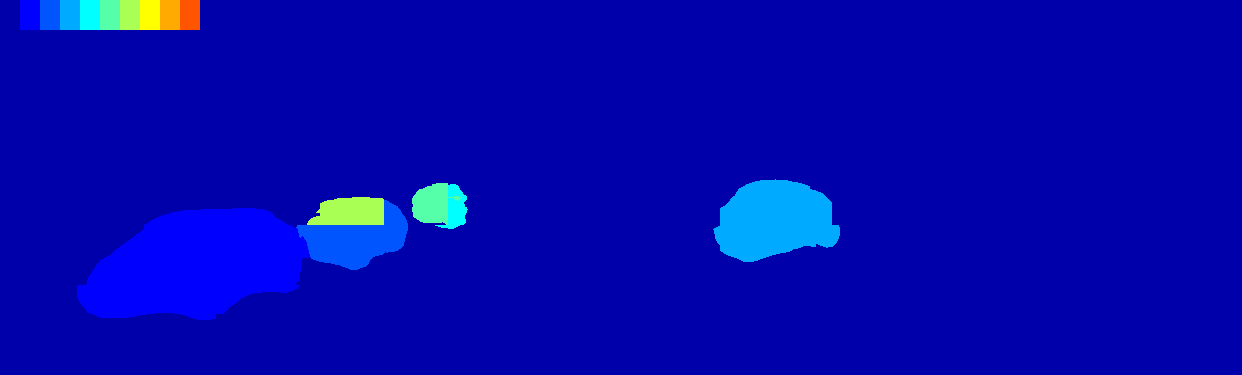}
\includegraphics[width=0.247\linewidth]{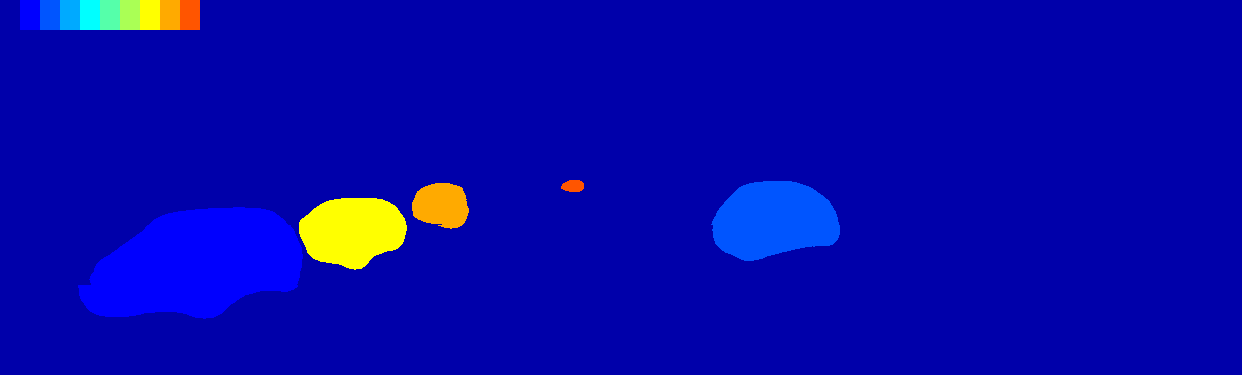}\\[0.3mm]
\includegraphics[width=0.247\linewidth]{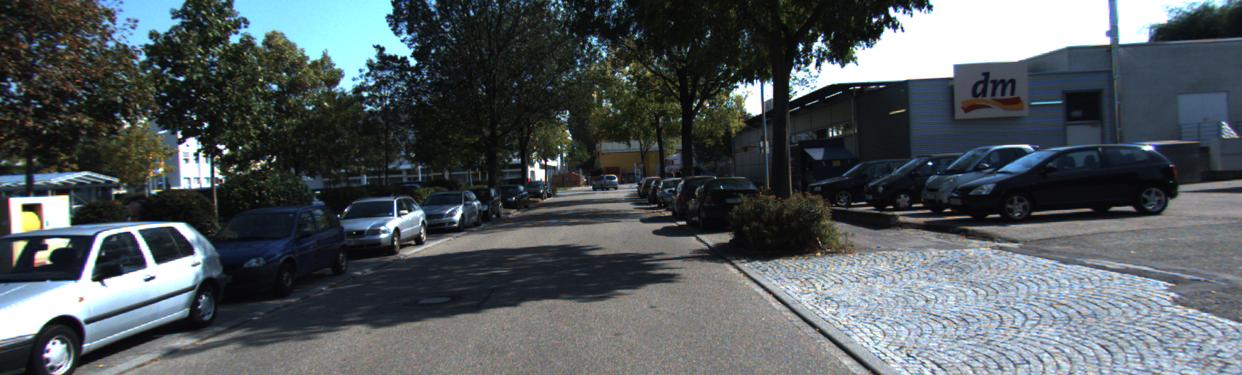}
\includegraphics[width=0.247\linewidth]{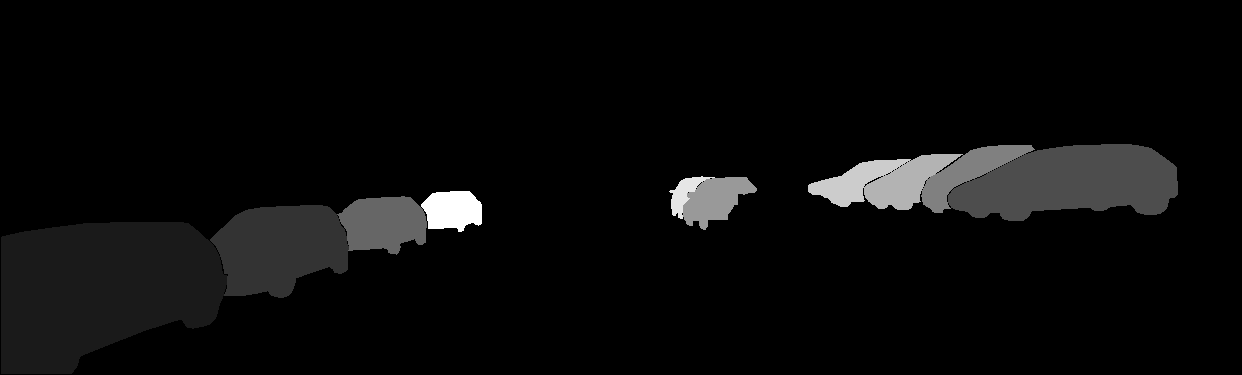}
\includegraphics[width=0.247\linewidth]{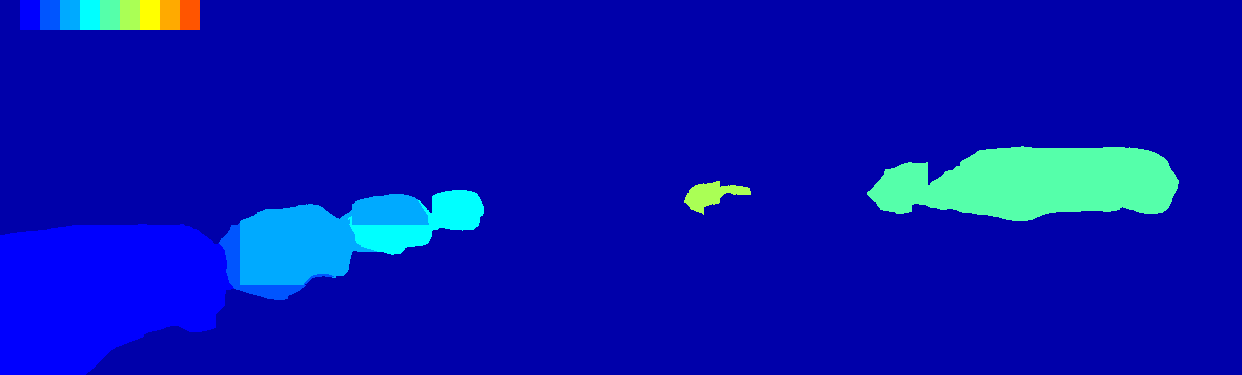}
\includegraphics[width=0.247\linewidth]{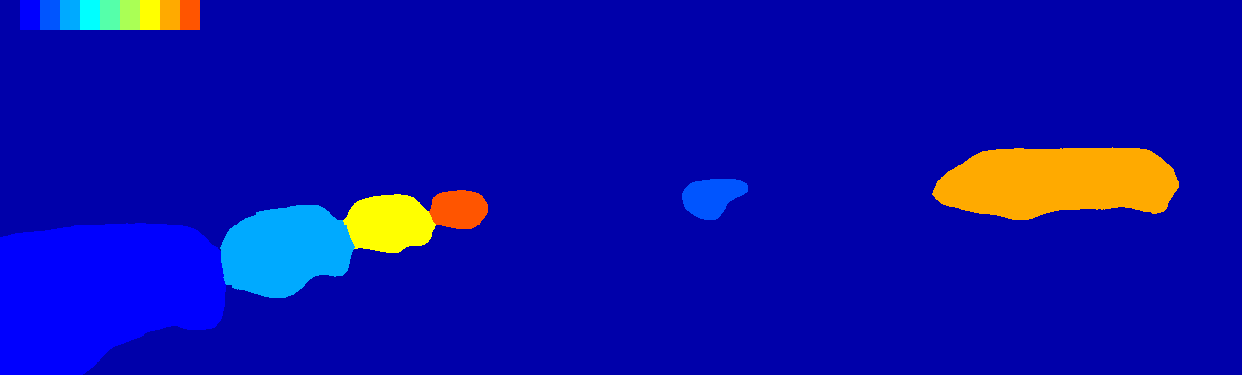}\\[0.3mm]
\includegraphics[width=0.247\linewidth]{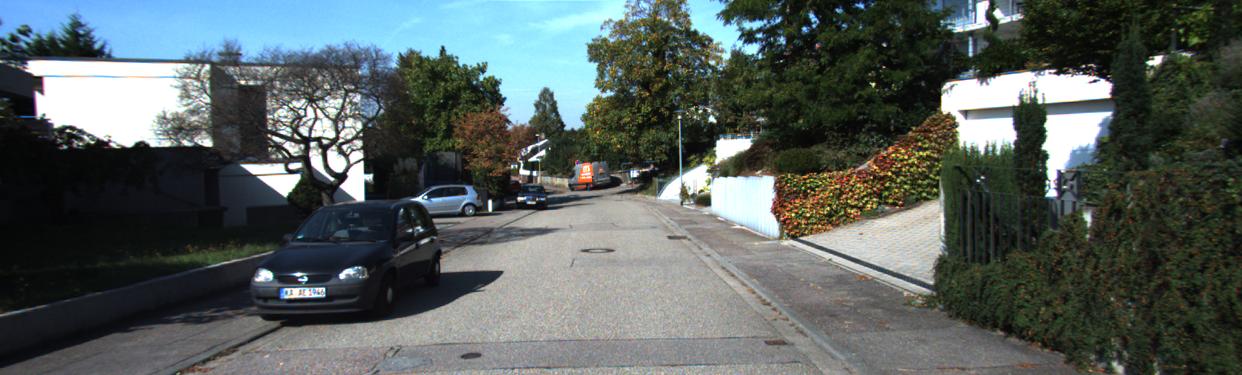}
\includegraphics[width=0.247\linewidth]{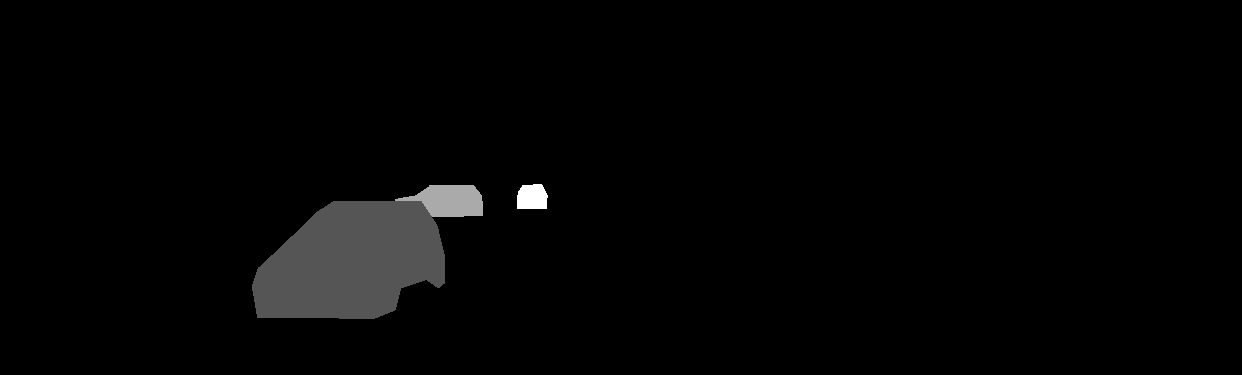}
\includegraphics[width=0.247\linewidth]{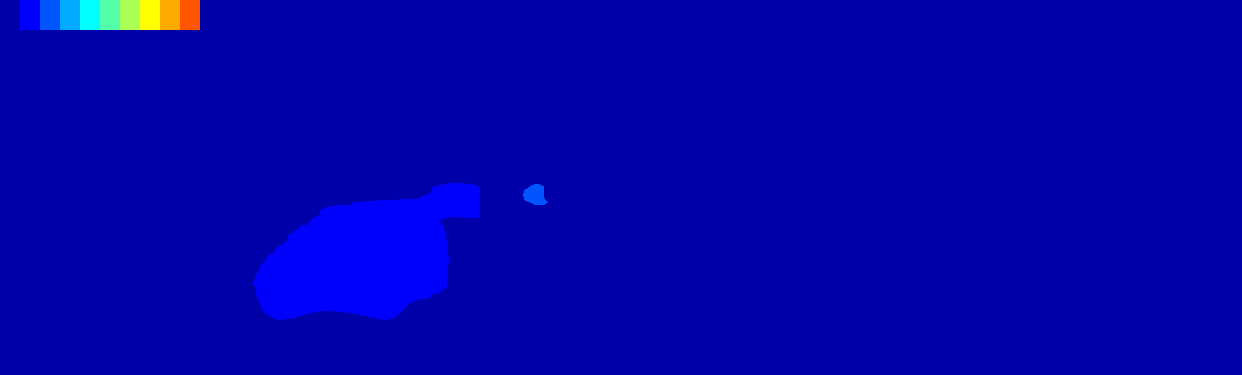}
\includegraphics[width=0.247\linewidth]{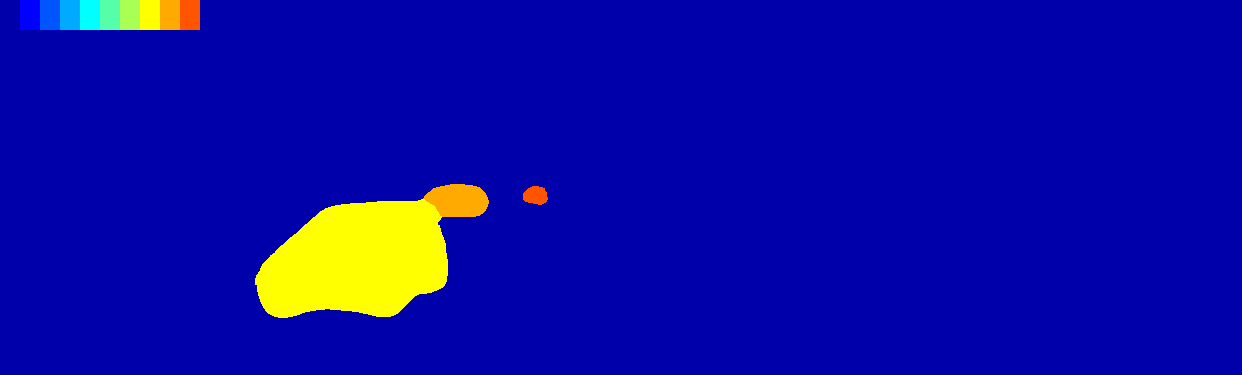}\\[0.3mm]
\includegraphics[width=0.247\linewidth]{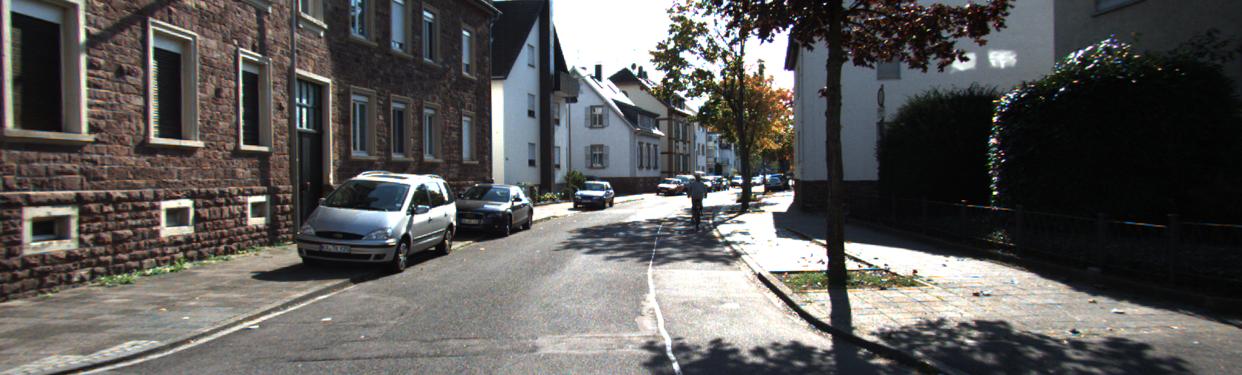}
\includegraphics[width=0.247\linewidth]{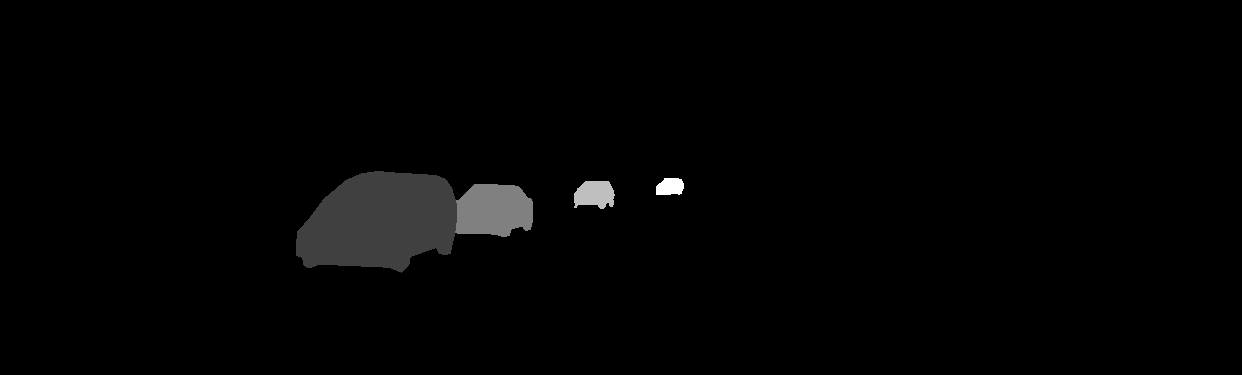}
\includegraphics[width=0.247\linewidth]{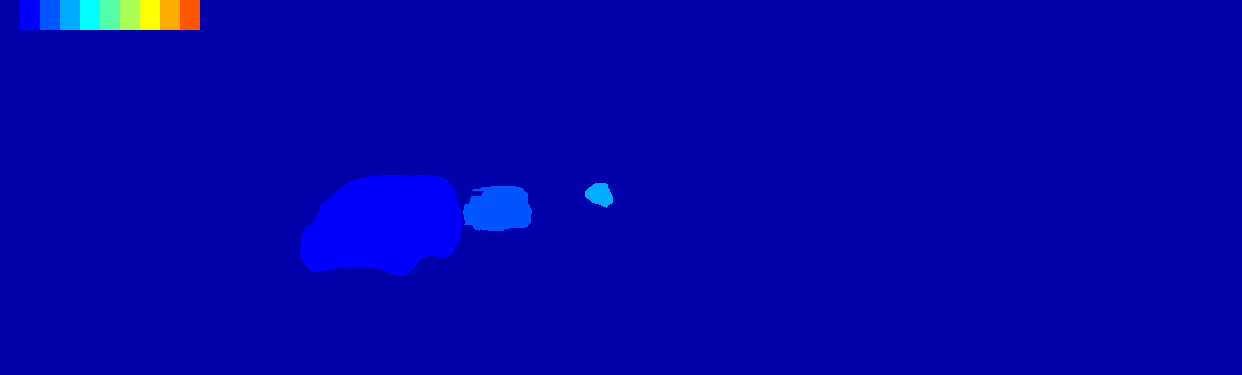}
\includegraphics[width=0.247\linewidth]{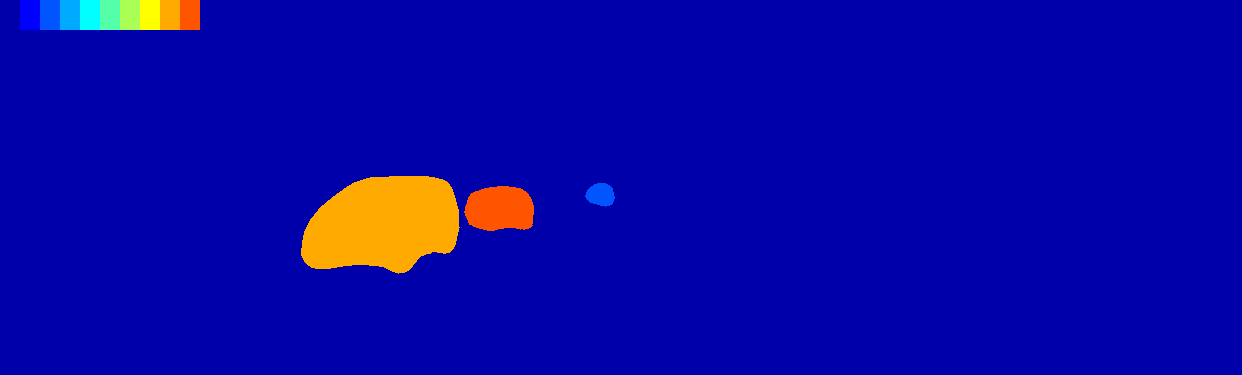}\\[0.3mm]
\includegraphics[width=0.247\linewidth]{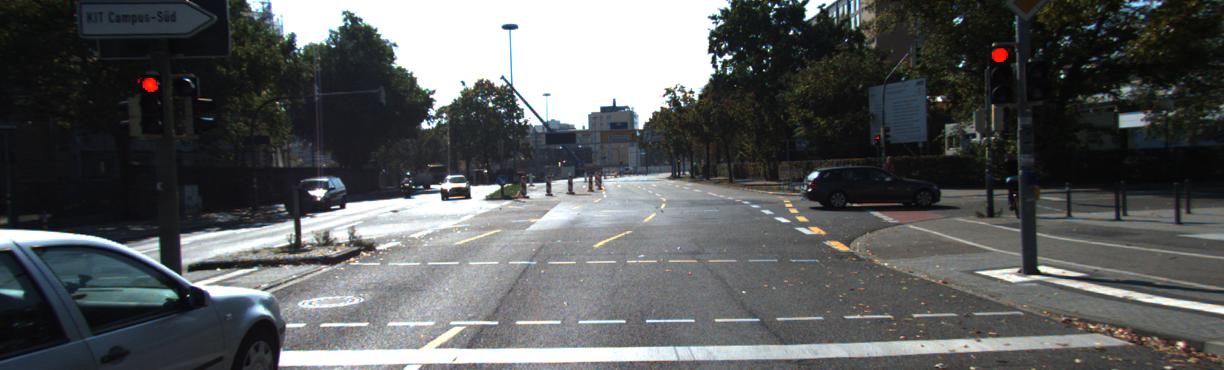}
\includegraphics[width=0.247\linewidth]{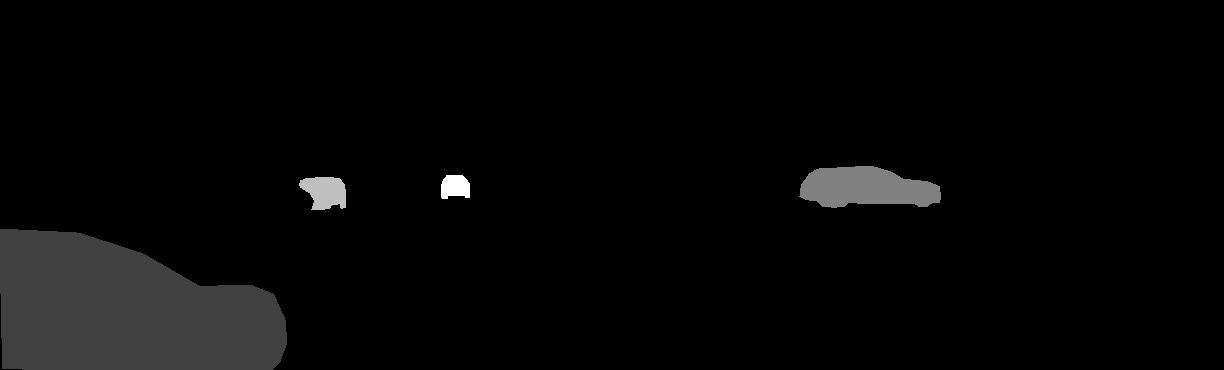}
\includegraphics[width=0.247\linewidth]{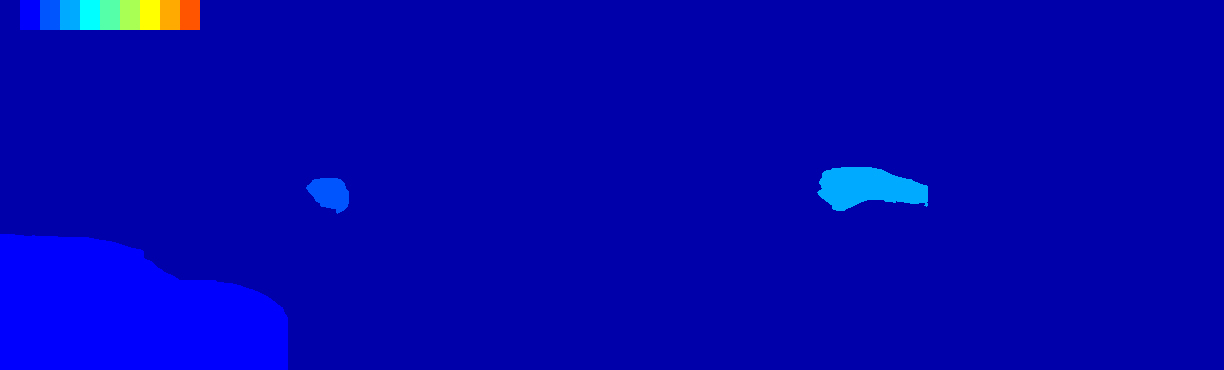}
\includegraphics[width=0.247\linewidth]{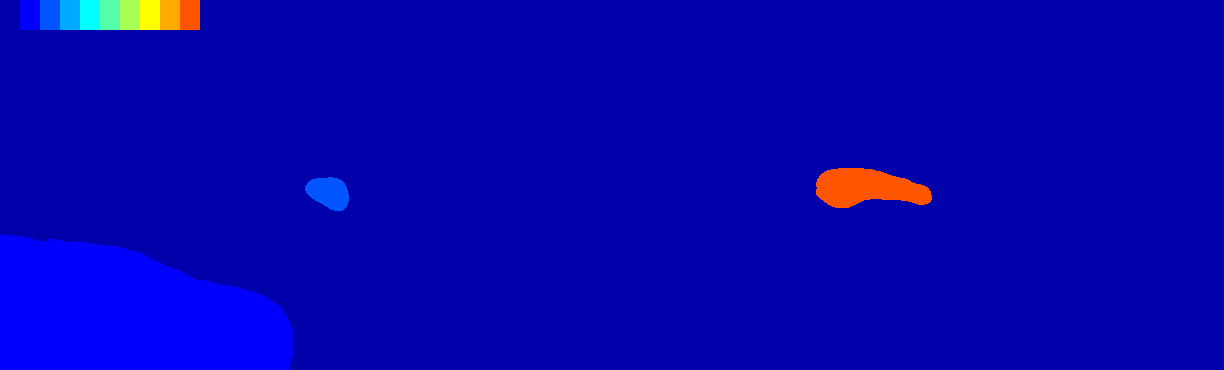}\\
\addtolength{\tabcolsep}{38.5pt}
\begin{tabular}{cccc}
Image & Ground Truth & ~\cite{ZhangICCV15}& $\ \ \ \ \ $Ours
\end{tabular}
\vspace{-3.5mm}
\caption{Successes of our model (without post-processing) with comparison to~\cite{ZhangICCV15}.}
\label{fig:success}
\vspace{-1mm}
\end{figure*}

\begin{figure*}[htb!]
\vspace{-1mm}
\includegraphics[width=0.247\linewidth]{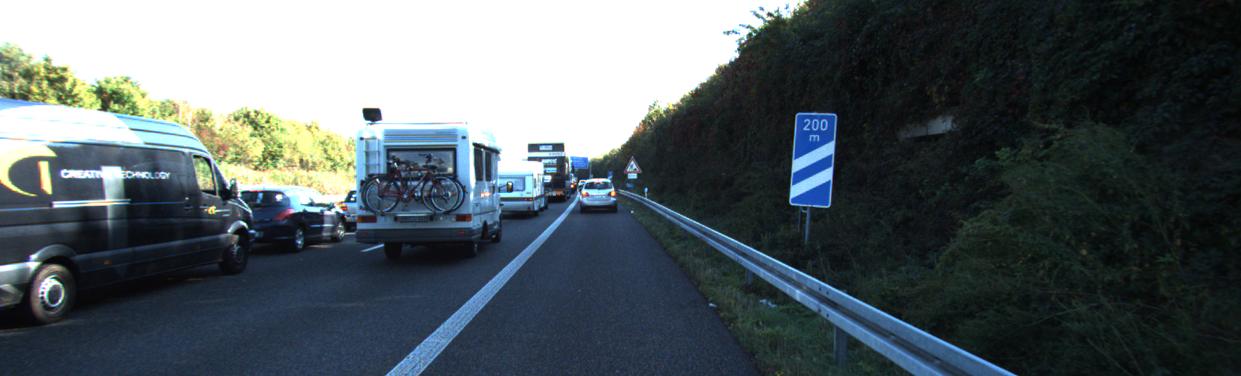}
\includegraphics[width=0.247\linewidth]{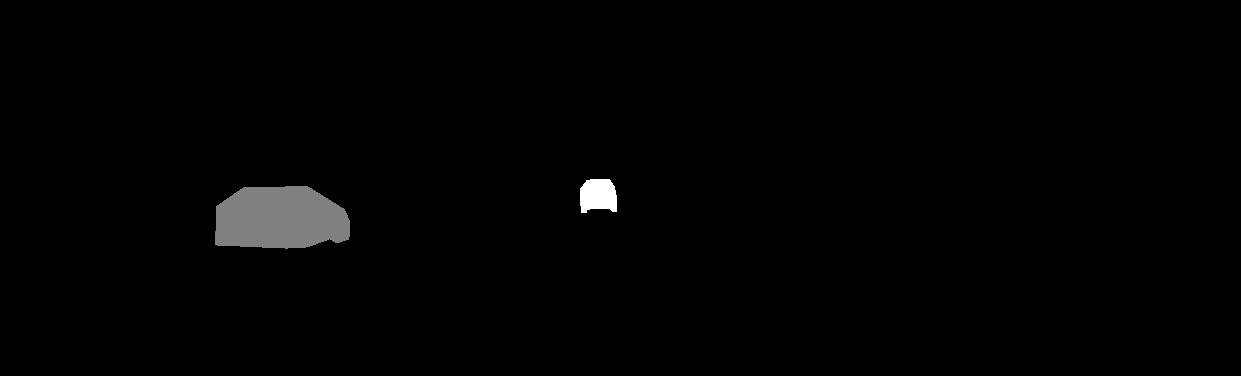}
\includegraphics[width=0.247\linewidth]{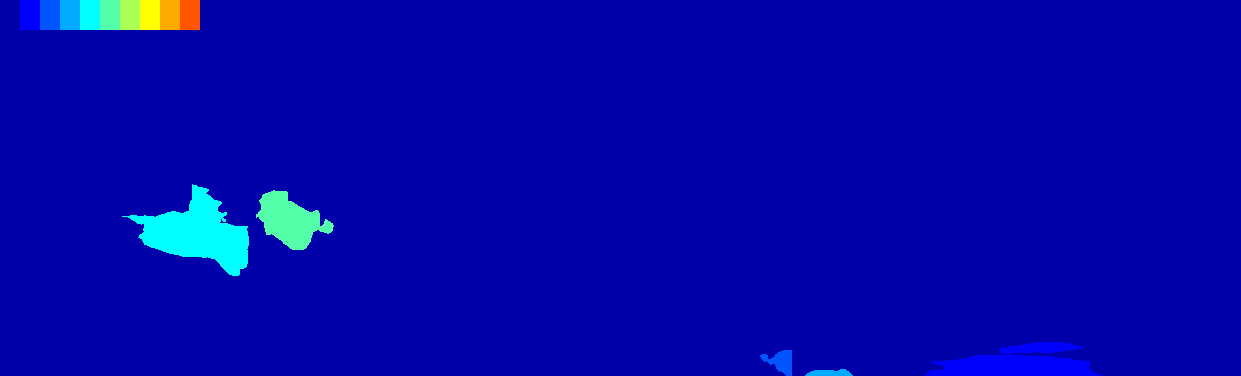}
\includegraphics[width=0.247\linewidth]{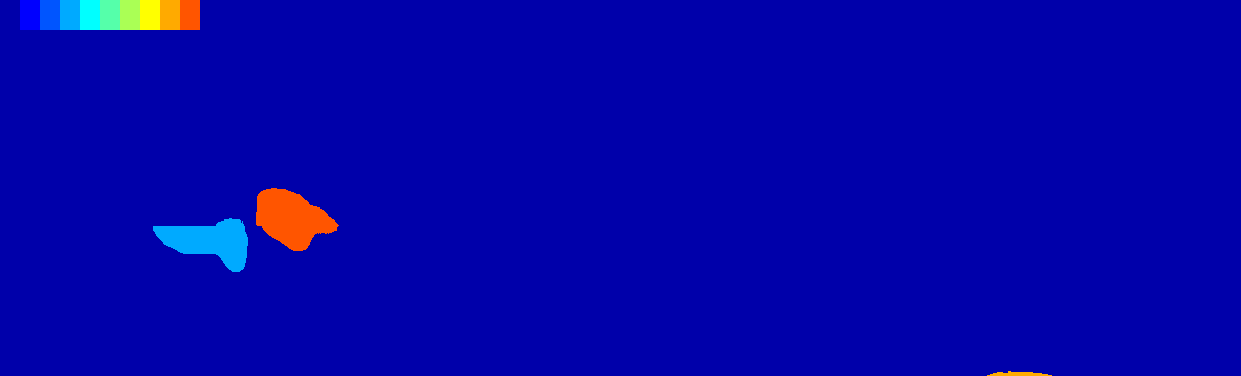}\\[0.3mm]
\includegraphics[width=0.247\linewidth]{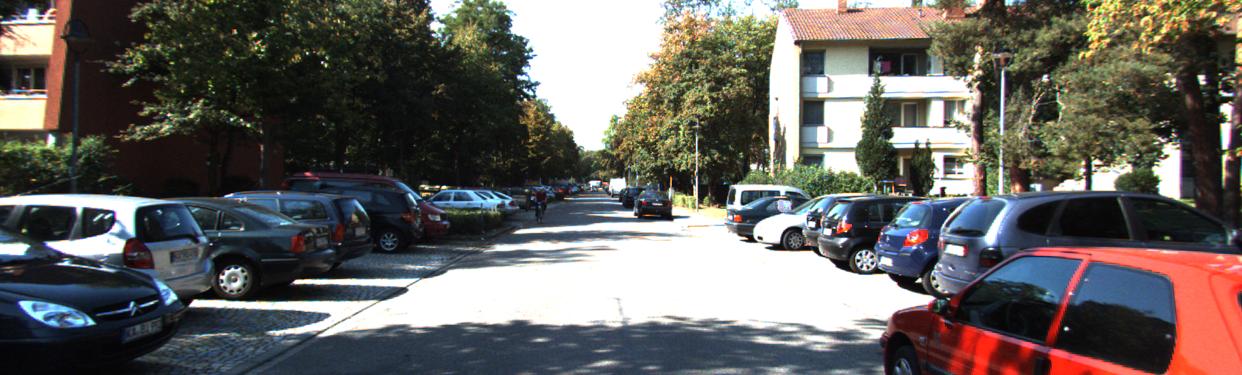}
\includegraphics[width=0.247\linewidth]{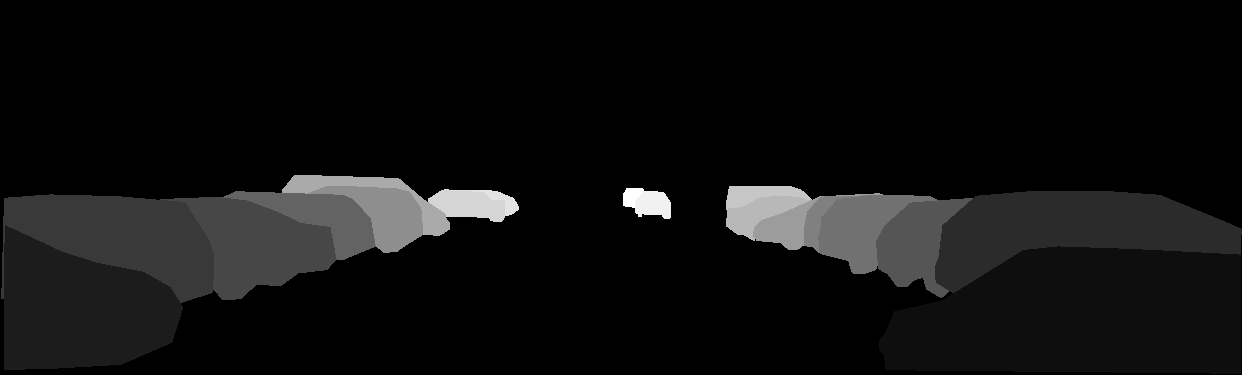}
\includegraphics[width=0.247\linewidth]{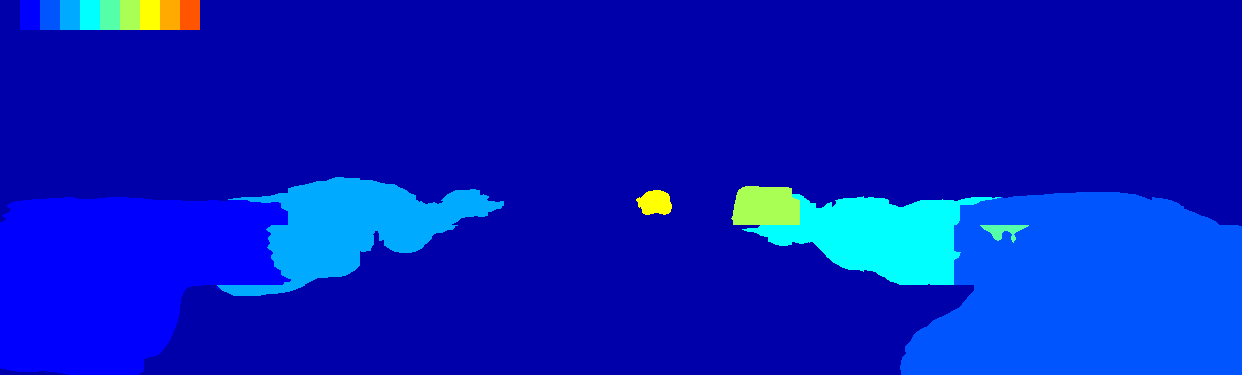}
\includegraphics[width=0.247\linewidth]{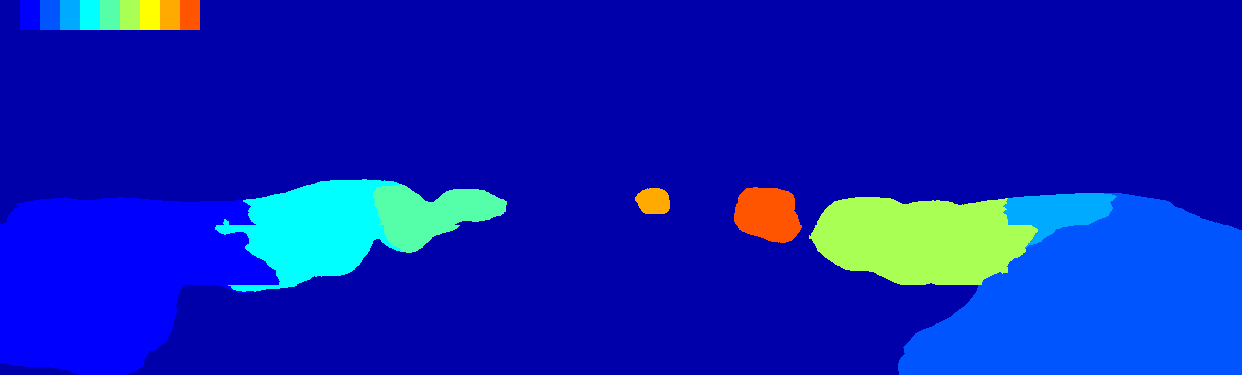}\\[-1mm]
\addtolength{\tabcolsep}{38.5pt}
\begin{tabular}{cccc}
Image & Ground Truth & ~\cite{ZhangICCV15}& $\ \ \ \ \ $Ours
\end{tabular}
\vspace{-3.5mm}
\caption{Failures of our model, mostly due to large non-car vehicles, small objects or severe occlusion.}
\label{fig:failure}
\vspace{-2mm}
\end{figure*}

\vspace{-0.4cm}
\paragraph{Evaluation Metrics.} Following~\cite{ZhangICCV15}, we use a number of metrics to provide a comprehensive evaluation of our model. We divide the metrics into two categories, namely class-level (i.e., \emph{Car} vs. \emph{non-Car}) and instance-level.
For the \emph{class-level evaluation}, we report the standard intersection-over-union score for the foreground `FIoU'. 

For the \emph{instance-level evaluation}, we provide the mean-weighted-coverage score and the mean-unweighted-coverage score introduced in~\cite{SilbermanECCV14}, which we denote by `MWCov' and `MUCov' respectively. For each ground-truth instance in a given image, we find its maximally overlapping prediction and compute the IoU score between them. The weighted-coverage score for the image is then the average of the IoU scores weighted by the size of the ground-truth instances. Finally, `MWCov' is obtained by averaging the weighted-coverage score across images. The mean-unweighted-coverage score is computed similarly but treats every ground-truth instance equally. `MWCov' and `MUCov' are  important, because they directly evaluate how closely our predictions overlap with ground-truth instances, since these two metrics are based on IoU scores. However, they do not penalize false positive instances.

For each predicted instance, we compute the ratio of class-level (\emph{Car} vs. \emph{non-Car}) true positive pixels inside it. The ratio is then averaged across predicted instances and reported as `AvgPr'. Similarly for each ground-truth instance, we compute the ratio of true positive pixels inside it. The ratio is then averaged across all ground-truth instances and reported as `AvgRe'. 

If a predicted instance does not overlap with any ground-truth instance, we deem it as a false positive instance. We average the number of false positive instances in each image across images, and report it as `AvgFP'. Similarly, if a ground-truth instance does not overlap with any prediction, we deem it as a false negative instance. We average the number of false negative instances in each image across images, and report it as `AvgFN'.

Finally, for each ground-truth instance, we find a prediction which overlaps more than 50\% with it. We divide the number of such GT-prediction pairs either by the number of predictions to obtain instance-level precision denoted by `InsPr', or by the number of ground-truth instances to obtain instance-level recall denoted by `InsRe', and report the corresponding F1 score denoted by `InsF1'. Intuitively, `InsPr' and `InsRe' reflect the model's ability to avoid false positive instances and false negative instances respectively at a 50\% threshold. The corresponding `InsF1' score unifies the previous two metrics. `InsF1' score (on the validation set) is the metric we use for selecting our model parameters.

\vspace{-0.4cm}
\paragraph{Quantitative Results.} The evaluation results on our test set are given in Tab.~\ref{tab:metricsTest}. We report results for  three instantiations of our model: `LocCNNPred' uses only the local CNN prediction term; `LocCNNPred+InterConnComp' adds the inter-connected component term; while `Full' denotes our full model. Following~\cite{ZhangICCV15}, we additionally apply a few post-processing steps including hole filling, removing tiny isolated regions and splitting the connected components of any prediction into separate instances. This is reported at the bottom of the table.

Notice that without post-processing the model `LocCNNPred' which has only the local CNN prediction term performs much worse than our baselines, because it allows instances that do not coexist in any patch to have the same labeling. 
Post-processing removes some of these mistakes and makes the model already outperform the baselines~\cite{ZhangICCV15} in a number of metrics.
With the addition of the inter-connected component term, the model `LocCNNPred+InterConnComp' encourages far apart instances to take different labels, which improves the performance. Finally, our full model which further adds the smoothness term is able to remove noisy regions scattered around the image, especially around instance boundaries where the CNN predictions are not confident. Our full model boosts instance-level precision by a huge margin compared to `LocCNNPred+InterConnComp', outperforming the baselines significantly in a number of metrics.


\vspace{-0.4cm}
\paragraph{Qualitative Results.} We show examples of successes of our model (`Full') without post-processing in \figref{fig:success}. 
We compare our full model to ground truth and the baseline `ConnComp' which has the highest `InsF1' score compared to the others.
While the baseline tends to merge neighboring instances into one, our model is more successful in telling them apart. Patch boundary is clearly noticeable in many of the baseline results, while our model prevents this from happening. This suggests that our model exploits the local CNN predictions in a much better way than the baselines.

A few failure cases of our model are shown in Fig.~\ref{fig:failure}, which are largely due to the CNN output. Vehicles like vans that are similar to cars (but not labeled as cars) tend to confuse the CNN, thus introducing false positives into our results. Heavily occluded cars also pose great challenges. 

\vspace{-2mm}
 \section{Conclusions}
\vspace{-1mm}

In this paper, we proposed a new approach for instance-level segmentation. Our approach builds upon the recently proposed work~\cite{ZhangICCV15} which trains a CNN on local patches to obtain soft instance labelings. We propose a densely connected MRF that is amenable to the efficient inference algorithm by~\cite{krahenbuhl2011efficient} to derive a globally consistent instance labeling of the full image. Our MRF exploits local CNN predictions, long-range connections between far apart instances, and contrast-sensitive smoothness. Our experiments show significant improvements over~\cite{ZhangICCV15}.

\vspace{0.2cm}
\noindent{\bf Acknowledgments:} This work was partially supported by ONR-N00014-14-1-0232, Samsung and NSERC.

{\small
\bibliographystyle{ieee}
\bibliography{egbib}
}

\end{document}